\documentclass[lettersize,journal]{IEEEtran}
\usepackage{amsmath,amsfonts}
\usepackage{algorithm}
\usepackage{algpseudocode}
\usepackage{array}
\usepackage[caption=false,font=normalsize,labelfont=sf,textfont=sf]{subfig}
\usepackage{textcomp}
\usepackage{stfloats}
\usepackage{url}
\usepackage{verbatim}
\usepackage{graphicx}
\usepackage{cite}
\usepackage{mathtools}
\usepackage{subfig}
\usepackage{multirow}
\usepackage{multicol}
\usepackage{xcolor}
\usepackage{colortbl}
\usepackage{nicefrac}
\usepackage{colortbl}
\begin{document}

\title{Shear-based Grasp Control for Multi-fingered Underactuated Tactile Robotic Hands}

\author{Christopher J. Ford, Haoran Li, Manuel G. Catalano, Matteo Bianchi, Efi Psomopoulou$^*$, Nathan F. Lepora$^*$
\thanks{CJF, EP and NL were supported by the Horizon Europe research and innovation program under grant agreement No. 101120823 (MANiBOT). NL and EP were also supported by the Royal Society International Collaboration Awards (South Korea). NL was also supported by an award from ARIA on `Democratising Hardware And Control For Robot Dexterity'.\newline
\indent CJF, HL, EP and NL are with the Department of Engineering Mathematics and Bristol Robotics Laboratory, University of Bristol, UK (e-mail: chris.ford@bristol.ac.uk, haoran.li@bristol.ac.uk, efi.psomopoulou@bristol.ac.uk, n.lepora@bristol.ac.uk). $^*$Equal contribution. \newline \indent MB is with the Department of Information Engineering and the Research Center "E. Piaggio", University of Pisa, Italy (e-mail: matteo.bianchi@centropiaggio.unipi.it). \newline \indent MGC is with the Department of Soft Robotics for Human Cooperation and Rehabilitation, Istituto Italiano di Tecnologia (IIT), Italy (e-mail: manuel.catalano@iit.it.)}}%

\markboth{IEEE Transactions on Robotics,~Vol.~XX, No.~X, XXXX}%
{Shell \MakeLowercase{\textit{et al.}}}


\maketitle

\begin{abstract}
This paper presents a shear-based control scheme for grasping and manipulating delicate objects with a Pisa/IIT anthropomorphic SoftHand equipped with soft biomimetic tactile sensors on all five fingertips. These `microTac' tactile sensors are miniature versions of the TacTip vision-based tactile sensor, and can extract precise contact geometry and force information at each fingertip for use as feedback into a controller to modulate the grasp while a held object is manipulated. Using a parallel processing pipeline, we asynchronously capture tactile images and predict contact pose and force from multiple tactile sensors. Consistent pose and force models across all sensors are developed using supervised deep learning with transfer learning techniques. We then develop a grasp control framework that uses contact force feedback from all fingertip sensors simultaneously, allowing the hand to safely handle delicate objects even under external disturbances. This control framework is applied to several grasp-manipulation experiments: first, retaining a flexible cup in a grasp without crushing it under changes in object weight; second, a pouring task where the center of mass of the cup changes dynamically; and third, a tactile-driven leader-follower task where a human guides a held object. These manipulation tasks demonstrate more human-like dexterity with underactuated robotic hands by using fast reflexive control from tactile sensing.
\end{abstract}


\section{Introduction}
In robotic manipulation, accurate force sensing is key to executing efficient, reliable grasping and manipulation without dropping or mishandling objects. This manipulation is particularly challenging when interacting with soft, delicate objects without damaging them, or under circumstances where the grasp is disturbed. For complex manipulation tasks, it is expected that multi-fingered dexterous manipulators with many degrees of freedom are needed \cite{bicchi2000robotic}. This freedom comes at the cost of greater complexity in their control, which can be partially ameliorated with the use of underactuated robot hands that have high degrees of freedom yet require far more basic control relative to their fully actuated counterparts, albeit at the expense of some dexterity~\cite{birglen2004kinetostatic, okamura2000overview}. Another aspect of force-sensitive manipulation, is to have detailed information on contact pose, normal force and shear force at the point of contact, which can be provided by high-resolution tactile feedback \cite{10161036}. The tactile feedback could also help compensate for the lower dexterity of underactuated manipulators, which is a viewpoint that will be explored in this paper.

An underappreciated component of robotic manipulation is shear sensing from the point of contact. A recent study using a similar tactile sensor as the present work, showed that complex interactions with objects such as continuous pushing or holding contact under translational/rotational motion are only possible with knowledge of the contact shear~\cite{lloyd2023pose}. While the grasp force may be inferred from the motor currents in fully actuated hands, this only resolves normal force. Furthermore, accurate measurements are not possible in underactuated hands without accurate kinaesthetic sensing, as grasp force cannot be accurately measured due to decoupling between the point of contact and actuators \cite{10161036}. Therefore, for soft underactuated robotic hands, suitable shear sensing at the point of contact is key to robotic manipulation. Yet not all modern tactile sensors integrated into robotic hands are appropriate for sensing shear, as many only detect normal force or have low resolution and/or sensitivity~\cite{kappassov2015tactile}.


A biomimetic marker-based tactile sensor, the Bristol Robotics Laboratory (BRL) TacTip, has been found particularly effective for slip and shear detection both with a single fingertip~\cite{james2018slip} and with multiple fingertips of a robot hand~\cite{james2020slip}.  The unique feature of the TacTip is its biomimetic morphology, with markers applied to the ends of pins rather than just the sensing surface of the skin, which is inspired by the sub-dermal papillae structures through which humans perceive touch~\cite{lepora2021soft, ward2018tactip}. Having the markers cantilevered in this way amplifies contact deformation, making the sensor highly sensitive to slippage and shear. Likewise, the TacTip can also use local shear information to detect incipient slip, facilitated by a fingerprint-like structure~\cite{james2020slip,bulens2023incipient}, although this remains a problem under investigation~\cite{chen2018tactile}. At the time of writing, whilst there has been progress in sensing shear force with tactile sensors, there has been no implementation of shear-based grasp control on a multi-fingered hand using feedback from multiple high-resolution tactile sensors. The benefit of this is that the sensors provide access to more information-rich contact data, which allows for more complex manipulation. The challenge comes from handling large amounts of high-resolution data, so that the processing does not slow down the system due to high computational demands.

In the present work, we propose a novel grasp control policy for underactuated hands where a state of `pre-slip' is maintained by balancing normal and shear contact forces, which we apply to the single-actuated anthropomorphic Pisa/IIT SoftHand~\cite{catalano2014adaptive}. For this control, we accurately predict three-dimensional contact pose and force at the point of contact from five tactile sensors mounted at the fingertips of the SoftHand using supervised deep learning techniques. The tactile sensors used are miniaturized TacTip optical tactile sensors (called `microTacs') developed for integration into the fingertips of this hand. This controller is applied to this underactuated grasp modulation during disturbances and manipulation. We perform several grasp-manipulation experiments to demonstrate the hand's extended capabilities for handling unknown objects with a stable grasp firm enough to retain objects under varied conditions, yet not exerting too much force as to damage them.

Our main contributions are:
\subsubsection{Control methods for handling objects with an underactuated tactile robotic hand} We present a novel grasp controller framework for an underactuated soft robot hand that allows it to stably grasp an object without applying excessive force, even in the presence of changing object mass and/or external disturbances. The controller uses marker-based high resolution tactile feedback sampled in parallel from the point of contact to resolve the contact poses and forces, allowing use of shear force measurements to perform force-sensitive grasping and manipulation tasks.

\subsubsection{Hardware advancements consisting of a complete tactile hand-arm robotics system with custom tactile fingertips and computational processing architecture} We designed and fabricated custom soft biomimetic optical tactile sensors called microTacs to integrate with the fingertips of the Pisa/IIT SoftHand. For rapid data capture and processing, we developed a novel computational hardware platform allowing for fast multi-input parallel image processing.

\subsubsection{Software methods for multi-tactile sensor learning of pose and shear}
A key aspect of achieving the desired tactile robotic control was the accurate prediction of shear and normal force and pose against the local surface of the object, for each tactile fingertip. This
was achieved using supervised deep learning methods extended to five tactile sensors concurrently (from one in previous work~\cite{lepora2021towards}). We find a combination of transfer learning and individual training gave the best models overall, as it allows for learned features from one sensor to be applied to the others.

\section{Background and Related Work} \label{sec:background}
\subsection{Force-sensitive grasping}
The elasticity of underactuated hands is beneficial for grasping performance, but introduces issues when considering force-sensitive manipulation. Historically, such applications employ some method to transduce the forces being experienced at each finger joint in order to resolve grasping force ~\cite{howe1993tactile, choi2012external, daoud2012real, nguyen2013fingertip, liu2015finger}. This method is straightforward when using rigid finger joints due to their predictable kinematics, but does not work well with soft underactuated hands and/or soft contacts due to the non-linearities introduced to the system from elasticity~\cite{grioli2012adaptive}. This is due to the elasticity in the kinematic chain absorbing an unknown amount of force from tha generated by the the payload mass, causing inaccuracies in inferring contact forces. This makes it difficult to apply other established methods for force sensing which require an analytical model of the system~\cite{barkat2009optimization, xu2021compliant}. 

It is well documented that measurements taken directly from the point of contact are essential when considering advanced robotic manipulation tasks, such as in-hand manipulation~\cite{howe1993tactile,tegin2005tactile, yousef2011tactile}. Ajoudani et al.\cite{ajoudani2016reflex} proposed a system where a Pisa/IIT SoftHand \cite{catalano2014adaptive} with tactile fingertips reacts to changes in grasped object mass or external disturbances on grasped objects, adjusting accordingly to retain the object. The fingertip sensors used were ThimbleSense sensors \cite{battaglia2015thimblesense} that are approximately the size of human fingertip, using an ATI Nano 17 Force/Torque sensor that could resolve 6-axis forces to a high degree of precision at a high frequency (0.003\,N resolution for translational force at 7\,kHz~\cite{ATI_Nano17}). However, they have limited sensory information compared with array-based or optical tactile sensors. For this reason, we focus here on using high-resolution soft optical tactile sensors.

\subsection{Tactile control}
Li et al.~\cite{li2013control} and Kappassov et al.~\cite{kappassov2020touch} present control frameworks for tactile servoing based on calculating geometric contact features suitable for guiding the sensor. Both studies use sensors with flat taxel arrays to infer contact features such as area of contact and the contact moments, which is then used as feedback into a servo-controller for the robot to execute contact tracking and contour-following tasks. These features are calculated analytically from the taxel array values, and are only suitable for control of rigid tactile arrays of known shape (e.g. flat in these two studies).

Work on tactile servo control with soft curved optical tactile sensors has instead used the surface contact pose as feedback to move a robot arm with tactile end effector  \cite{lepora2021pose}. The components of the surface contact pose are found by training  a convolutional neural network (CNN) to predict the contact pose through supervised deep learning over a dataset of tactile images of known poses~\cite{lepora2020optimal}. In the original work, shear was treated as a `nuisance variable' and the training data collection introduced random unlabelled shear perturbations, which enabled the learnt model to be insensitive to unknown motion-dependent shear. Later work has extended the model to also predict shear displacement~\cite{lloyd2023pose}, but required use of technically-sophisticated methods involving a Bayesian filter and Gaussian density network to reduce prediction uncertainties, for example from slippage during training data collection.

Shear cannot be ignored when handling an object against the force of gravity, as it provides key information on how to secure the held object \cite{de2017multimodal}. Meanwhile, tactile servo control without shear has been successfully applied to many single-sensor tasks, such as contour following~\cite{lepora2019pixels}, surface following~\cite{lepora2021pose} or pushing tasks \cite{lloyd2021goal}; however, these tasks only require measurements of contact depth and orientation to achieve. Lloyd and Lepora introduced pose and shear-based tactile servoing able to predict shear displacement~\cite{lloyd2023pose} as described above, however the Bayesian filtering required an estimate of the changes in tactile sensor pose, given by the robot arm kinematics on which the sensor was mounted. Thus, their approach cannot be applied to underactuated robot hands without knowledge of the finger joint positions. 

In the present work, we use a force/torque (FT) sensor to sample the resultant normal and shear contact force as part of the data collection protocol and so train a model to predict force. As stated, previous work in predicting 6-degree of freedom (DoF) pose with the TacTip shows that shear displacement is difficult to predict accurately due to the sensor slipping during data collection~\cite{lloyd2023pose}. This introduces discrepancies between the pose label and the resulting tactile image, which manifests as noise in the data. However, by sampling the resulting shear force from a given shear displacement we generate the label \textit{after} any slip event has occurred. This effectively filters out noise in the data due to slip, enabling accurate shear-force predictions.

\subsection{Robot Hands}
Robots intended for manipulation tasks should provide dexterity while minimizing the complexity of their control, which requires a balance between the degrees of freedom and degrees of actuation (DoAs), which is a quantity that defines the number of motors in the hand. For example, a fully-actuated hand with many DoFs such as the Shadow dexterous hand is very dexterous, but can be complex to control~\cite{andrychowicz2020learning}. Reducing the DoAs to simplify control is one solution, but does sacrifice some dexterity \cite{cabas2006optimized,deimel2016novel,li2022brl}. One way to compensate for this loss of dexterity is to use compliance in the robot hand~\cite{shintake2018soft}, allowing the hand to passively conform to the shape of held objects~\cite{niehues2015compliance}.

Here we use the Pisa/IIT SoftHand: an anthropomorphic hand that modulates the grasp in a whole hand pinch grasp, based on a motion derived from human postural synergies~\cite{catalano2014adaptive}\cite{della2018toward}. This hand is highly underactuated, with only 1 DoA, yet 19 DoF. It would be difficult to achieve good grasping performance with such limited actuation; however, the SoftHand mitigates this loss of dexterity (albeit at the loss of joint proprioception and therefore understanding of the exact grasp pose) by adding compliance into the mechanism via `dislocatable' joints in the fingers, realised as rigid phalanxes coupled via elastic linkages with a tendon-drive mechanism~\cite{della2018toward}. Together with the human-like postural synergy of grasp closure this forms an `adaptive synergy' where grasping performance is improved despite a lack of dexterity from underactuation~\cite{catalano2014adaptive}. 

\section{Hardware Methodology}
\subsection{The microTac soft biomimetic optical tactile sensor}
\begin{figure}[t!]
        \centering
        \includegraphics[width = 0.475\textwidth]{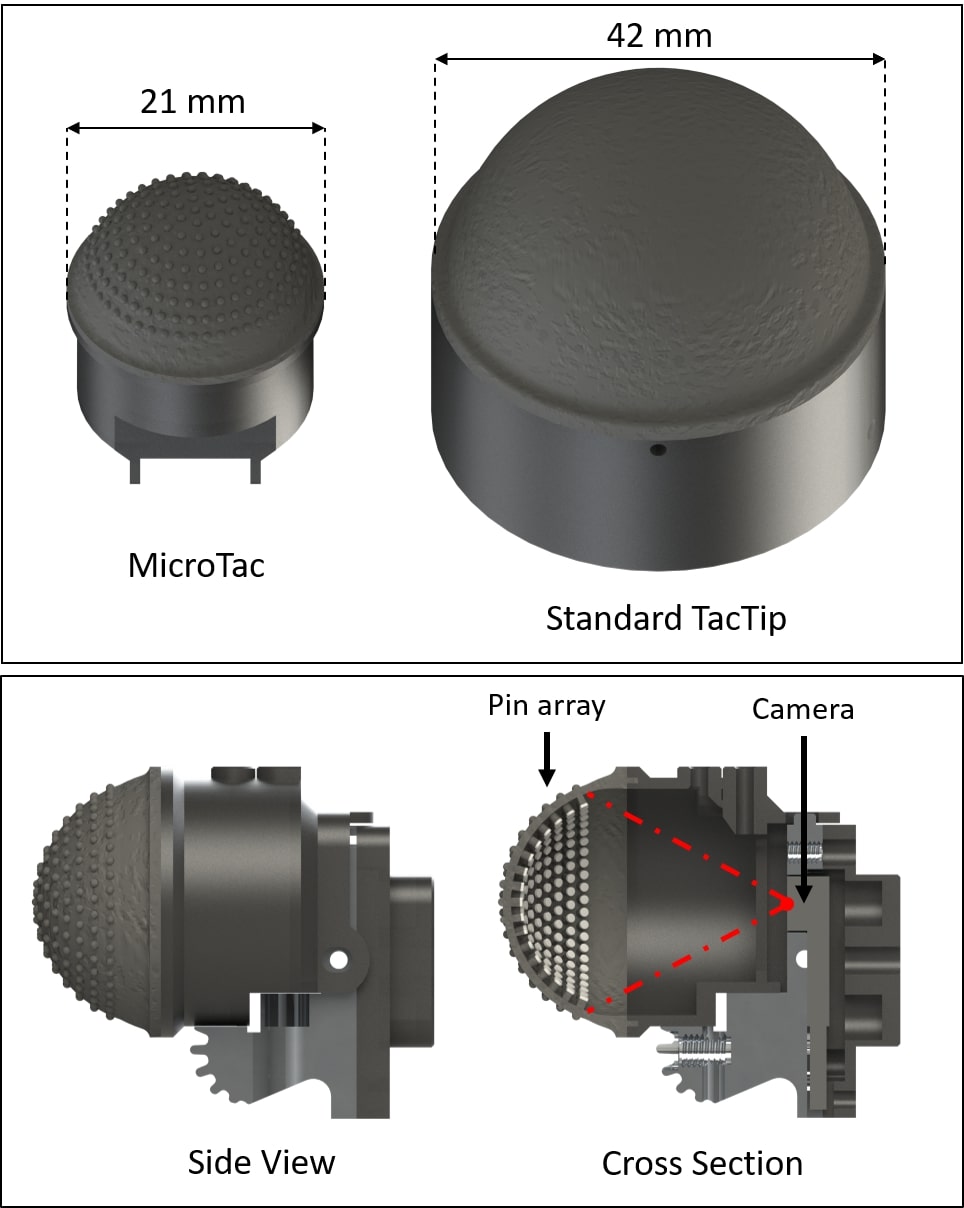}
        \caption{CAD model of the microTac soft biomimetic optical tactile sensor. (Top) The microTac dome compared to a standard TacTip dome. (Bottom) Side and cross-sectional views of the microTac, showing the internal camera viewing the marker-pin array protruding into the fingerprint bumps. Overall, the microTac is approximately the same size as the pad of a human fingertip.}
        \label{fig:tactip}
\end{figure}

\subsubsection{Maximizing marker density}
As we will require accurate shear information on human-fingertip sized sensors, we considered it important to maximize the number of pins and markers (the papillae-like structures on the interior of the skin and the spots of contrasting color to the skin, located on the tips of the pins, respectively). Markers are the primary features for predicting shear, so higher marker density should improve sensor prediction accuracy~\cite{ward2018tactip, li2023marker}.

It should be noted that although images are down-sampled, marker density and down-sampling are independent factors influencing performance. Down-sampling trades off some benefits of high marker density but would have more impact on sensors with low marker density~\cite{wang2021gelsight}. 

Improving the marker resolution is particularly important when considering a smaller sensor due to the inherently smaller visible contact area. However, this proves challenging for a TacTip, as smaller pins and markers can break more easily during manufacture. To address this issue, we developed a design approach to parametrically alter the number of pins as well as their diameter and height. This resulted in an efficient iterative workflow through which we were able to quickly determine the highest marker resolution able to be manufactured, to create a new sensor designed specifically as a fingertip for the Pisa/IIT SoftHand that we call the \textit{microTac} (design shown in Fig.~\ref{fig:tactip}; comparison with TacTip shown in Table \ref{tab:design}). This design study resulted in a sensor with 217 pins (from 127 pins on a larger TacTip) despite reducing the diameter of the sensor by 50\% from 4.2\,cm to 2.1\,cm. In terms of marker density, this translates to approximately 31 markers/cm$^2$ as opposed to 4 markers/cm$^2$ on a full-sized TacTip, giving an increase of marker resolution by 775\%. 

\begin{table}[t!]
\centering

\caption{Proposed microTac improved resolution compared to the standard TacTip from Refs~\cite{james2018slip,lepora2021pose,lepora2021soft} and DigiTac~\cite{lepora2022digitac}.}
\begin{tabular}{|c|c|c|c|}
\hline
               & Standard TacTip & DigiTac &\textbf{microTac} \\ \hline
Dimensions (cm)  & 4.2 dia.             & 2.5$\times$ 1.9 &\textbf{2.1} dia.             \\
No. of Pins    & 127             & 140 &\textbf{217}              \\
Markers/cm$^2$ & 4               & 16 &\textbf{31}                \\ \hline
\end{tabular}
\label{tab:design}
\end{table}

Table~\ref{tab:design} also shows a comparison with another sensor developed in Bristol Robotics Laboratory, the \textit{DigiTac}~\cite{lepora2022digitac}, which is relevant as it has also been integrated into the fingertips of robot hands~\cite{lu2024dexitac, yang2024anyrotate}. It is a version of the Meta DIGIT~\cite{lambeta2020digit} modified to sense tactile information through the TacTip pins-and-markers mechanism. Table~\ref{tab:design} shows the microTac is smaller than the DigiTac, yet has a pin density almost 50\% larger.

As far as we are aware, this design represents the highest marker density on this type of optical tactile sensor at the time of writing. Other works describe high-density marker arrays~\cite{abad2020visuotactile}; however, those tactile sensors feature 2D marker patterns printed on the skin or embedded in the elastomer. As discussed in the introduction, the markers on the TacTip skin are cantilevered on the end of pins in an artificial papillae structure, which increases the sensitivity by amplifying deformation of the skin into shear displacement of the markers. This TacTip design requires a morphologically-complex skin structure that is enabled by modern 3D-printing techniques, but nevertheless makes it more challenging to optimize aspects such as marker density. 

The sensor is printed as a single part on a Stratasys J826 Prime multi-material 3D printer, with the tip body made of Vero Black, the skin of Agilus30 Black and the markers of Vero White. After printing, a clear acrylic window is installed to form a sealed cavity beneath the skin, into which TECHSiL RTV27905 clear silicone encapsulant is injected.

\subsubsection{Artificial Fingerprint}
The microTac features an artificial fingerprint, based on a patterns of bumps that have previously been shown to improve the spatial acuity of TacTip-style sensors \cite{cramphorn2017addition}. Here this outer skin structure was primarily used to extend the effective pin length, as the height of the pins was reduced to accommodate the higher marker density. The fingerprint structure (cross-section in Fig.~\ref{fig:tactip}) is comprised of many raised sections of outer-skin material, each of which is concentric with a pin on the interior-skin. 

\subsection{Robotic tactile hand-arm and computing system}

\begin{figure}[t!]
    \centering
    \begin{tabular}{@{}c@{}c@{}}
        \begin{minipage}[b]{0.475\textwidth}
            \centering
            \small{\bf (a) Robot arm-hand system}\\
            \includegraphics[width=\textwidth]{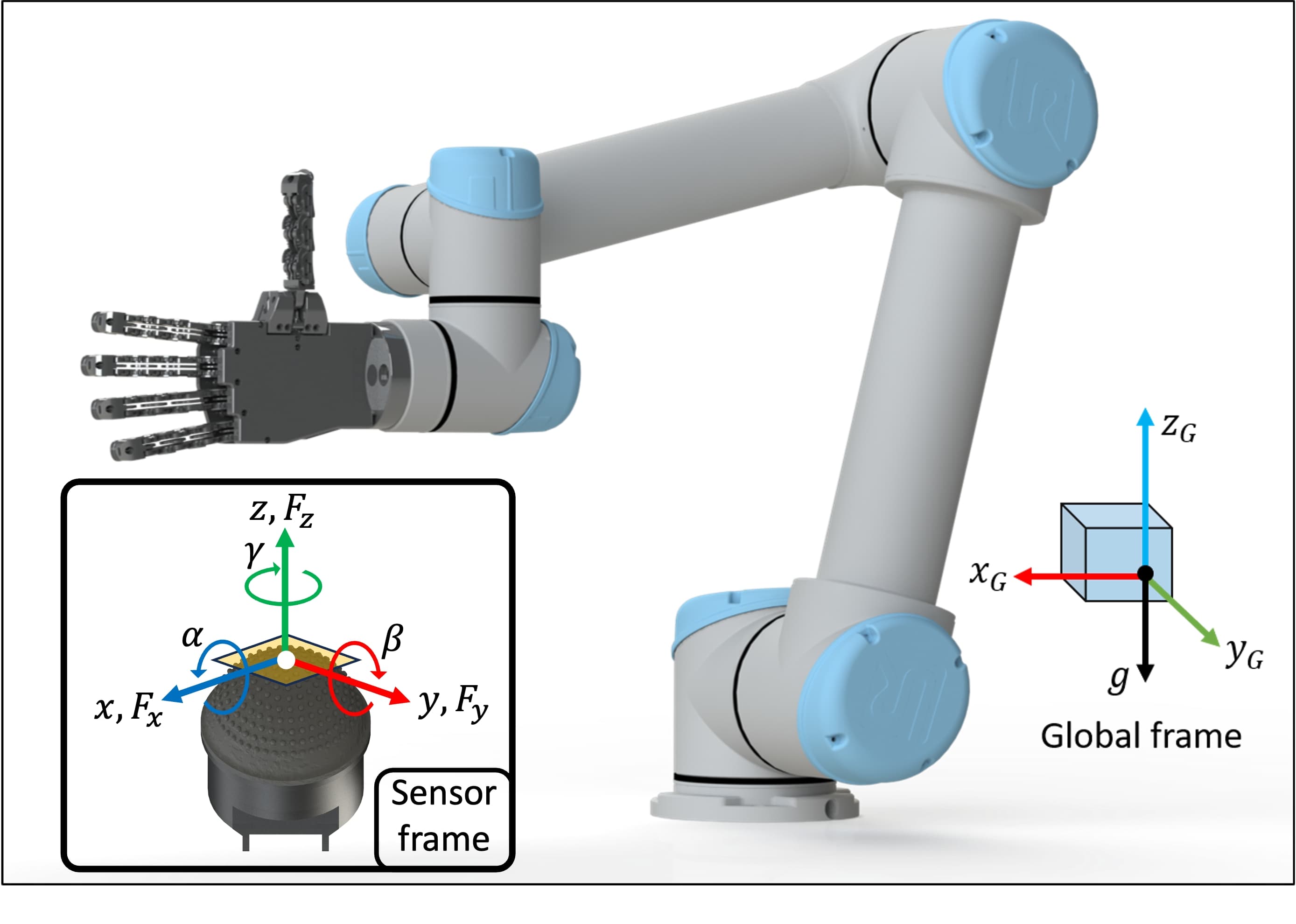}
        \end{minipage} \\
        \begin{minipage}[b]{0.475\textwidth}
            \centering
            \small{\bf (b) Computing infrastructure for tactile processing and control}\\
            \includegraphics[width=\textwidth]{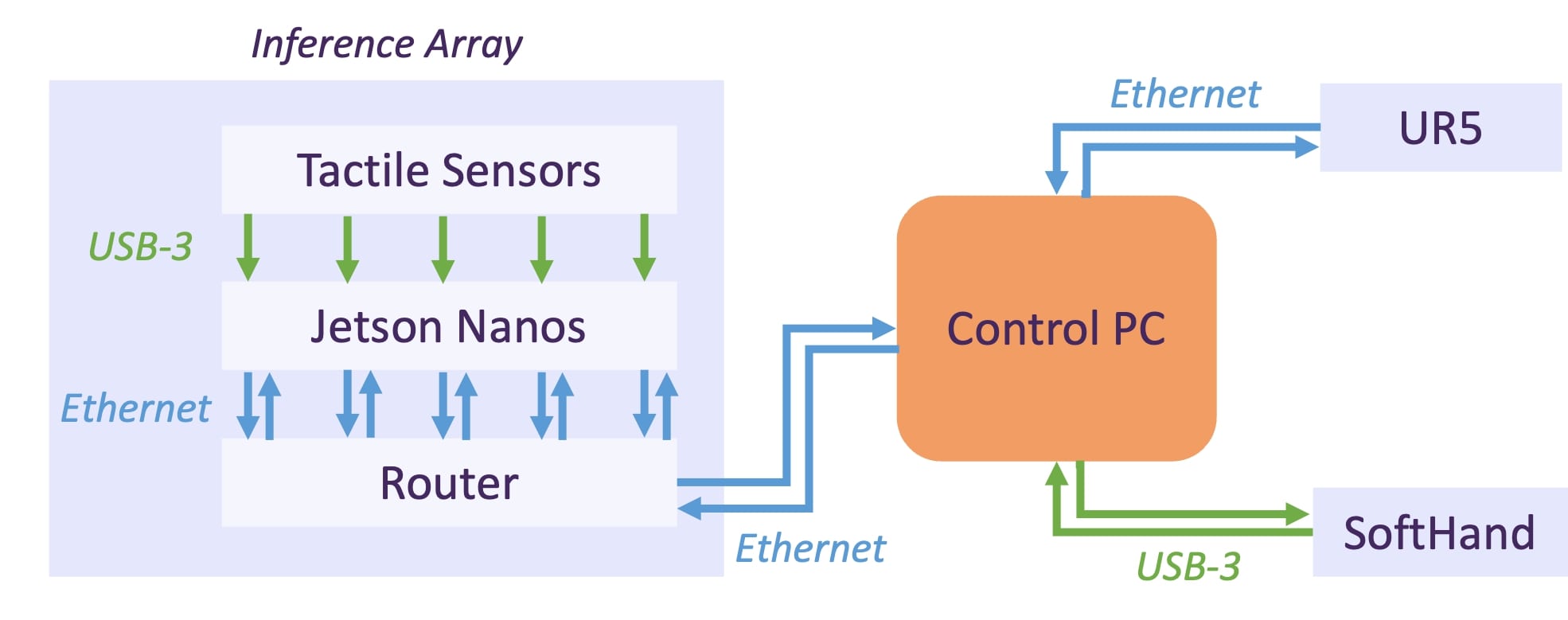}
        \end{minipage} \\
        \begin{minipage}[b]{0.475\textwidth}
            \centering
            \small{\bf (c) Image processing pipeline}\\
            \includegraphics[width=\textwidth]{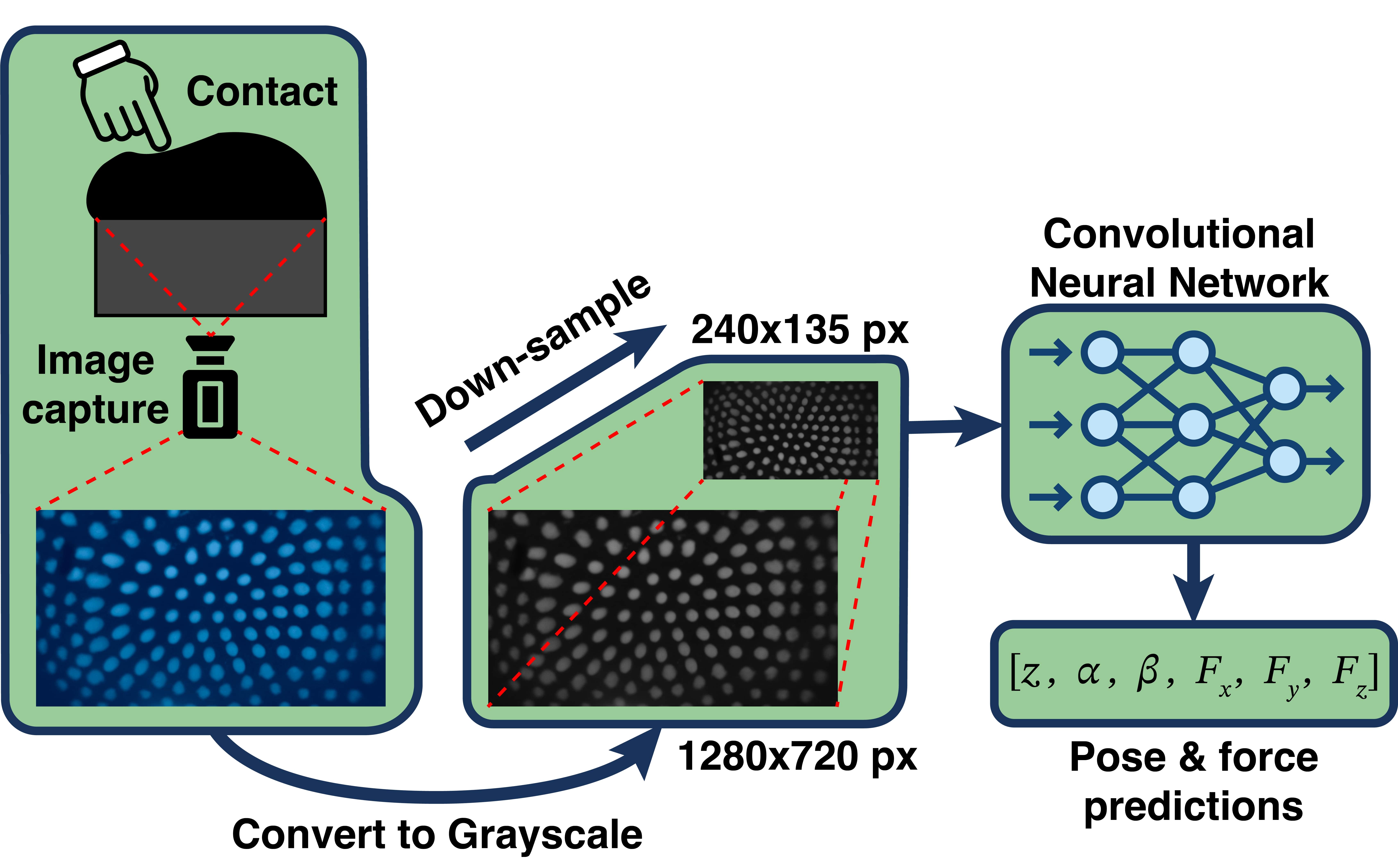}
        \end{minipage}
    \end{tabular}
    \caption{Robot hardware and computing infrastructure. (a) The microTac tactile fingertips mounted on the Pisa/IIT SoftHand, which is mounted on the wrist of a UR5 robot arm. The arm-hand system is shown oriented in the neutral pose in which the palm faces horizontally with the thumb up. Also shown is the sensor frame, illustrating how pose and force variables are oriented relative to the sensing surface. (b) The system architecture of the sensing and computing components of the robotic system, comprising tactile sensor inputs to a Jetson Nano array, coupled by a router to a control PC which controls the UR5 robot arm and Pisa/IIT SoftHand. The Jetson Nano array allows on-board tactile image capture, processing and model prediction, to minimize the computational load on the control PC. (c) A visualization of the image processing pipeline, illustrating how skin deformation from sensor contact is imaged, processed and converted to pose and force values.}
    \label{fig:robo_platform}
\end{figure}

The experimental robot system we use in this work (Fig.~\ref{fig:robo_platform}) is comprised of two actuating robots: a Universal Robots UR5 6-DoF industrial robot arm and a 1-DoA Pisa/IIT anthropomorphic SoftHand, as well as a tactile sensing component comprising the five microTac tactile sensors. The tactile sensors are mounted as fingertips on the SoftHand, which is in turn mounted as an end-effector on the UR5 robot arm. Each system element is connected to a central control PC (Dell Precision with 16GB RAM and Intel Core i7-13700), to process the tactile information and control the robot and hand. Communication with the actuating elements is via their proprietary software APIs in a way that allows for asynchronous control of the robot arm and SoftHand.

A new challenge was posed by capturing five USB camera-based tactile sensor inputs simultaneously at minimal latency and a reasonable frame rate, as most standard PCs do not have sufficient processing buses. Therefore, we developed a novel processing architecture that can capture and process tactile images from each sensor in parallel. This consists of five Nvidia Jetson Nano ARM64 Micro-computers (one for each sensor) which capture tactile images at 1280x720p, then process and down-sample the images to 240x135p (using bilinear interpolation) before passing through a neural network model processing pipeline illustrated in Fig~\ref{fig:robo_platform}c. These low-cost Jetson boards each has a solid-state GPU with 128 CUDA cores, which makes them capable in deploying embedded AI applications (as opposed to other micro-computers, such as a Raspberry Pi). The Jetson array means that data from the sensors is captured and processed in parallel, which gives a significant benefit over managing separate processes or threads on the control PC, as the computational demands associated with capturing and processing high-resolution tactile images are distributed across multiple sub-modules. We found this parallel processing to be essential, since images need to be captured at high resolution with low latency to process the relevant tactile information for the robot control. Utilizing the Jetson array means we can capture images from all sensors in parallel at a high resolution, with subsequent image processing executed asynchronously on each board. \textcolor{black}{This parallel image capture reduces latency that would be introduced by capturing images sequentially. Consequently, the sensing frequency is limited only by the camera frame rate, allowing us to run our control algorithm with an average sampling rate of 60\,Hz.} 

Data from the sensors is accessed using the Pyro4 distributed computing library to instantiate virtual sensor objects on the local network, which gives a class of functions for capturing images, processing images and running model inference. Using Pyro4, these functions can be accessed virtually from the control PC and used in a central Python control program as if the sensors were connected to the PC directly. This process reduces the size of the data packets coming to the control PC from the tactile sensors, so they can be managed in separate threads (along with communications to and from the UR5 arm and Pisa/IIT SoftHand), resulting in an asynchronous robotic system capable of real-time response and control.

\section{Software Methodology}
\subsection{Contact Pose and Force Data Collection} \label{sec:data_collect}

\begin{table}[b!]
\centering
\caption{Pose and shear parameter ranges used for training data collection, and corresponding ranges of the measured force.}
    \begin{tabular}{|c|c|c|c|}
    \hline
    \textbf{Pose and shear} & \textbf{Ranges}  & \textbf{Force} & \textbf{Ranges}\\ \hline
    $x$-shear                & $[-2, 2]$\,mm       & $F_x$        & $[-4,4]$\,N\\
    $y$-shear                & $[-2, 2]$\,mm       & $F_y$        & $[-4,4]$\,N\\
    $z$                & $[0,4]$\,mm         & $F_z$         & $[0,12]$\,N\\
    $\alpha$           & $[-20^\circ, 20^\circ]$ & $\tau_x$      & -\\
    $\beta$            & $[-20^\circ, 20^\circ]$ & $\tau_y$      & -\\
    $\gamma$-shear           & -                 & $\tau_z$      & -\\ \hline
    \end{tabular}
    \label{tab:pose_ranges}
\end{table}

The tactile robot was used in a different configuration to collect data for training. A custom end effector mount was fabricated so a microTac tactile sensor could be mounted on the wrist of the UR5 robot arm. In addition, a force-sensitive surface was created comprising a flat surface mounted on an ATI Mini 40 Force/Torque sensor (which was chosen for its high accuracy, low noise and low drift), which was itself mounted in a fixed position relative to the robot arm. 

The microTac was brought into contact with the force-sensitive surface at a variety of pre-determined poses sampled randomly from uniform distributions within the ranges in Table~\ref{tab:pose_ranges}. The data collection is based on that used to `train accurate pose models for surfaces and edges' from Ref.~\cite{lepora2020optimal}, which formed the basis for prior work on tactile servo control with the TacTip~\cite{lepora2021pose,lloyd2021goal,lloyd2023pose}. The surface pose relative to the tactile sensor is set by the contact depth and orientations $(z,\alpha,\beta)$, and the contact shears $(x,y)$ by an additional shearing motion while the sensor is in contact with the surface. These pose and shear parameters result in a normal and shear contact force $(F_x,F_y,F_z)$ whose measured ranges are also shown in Table~\ref{tab:pose_ranges}. 

Rotational $\gamma$-shear and related torques were omitted as preliminary testing indicated that torsional slippage occurred at around \textcolor{black}{1.3\,Nmm} corresponding to \mbox{$z$-displacements} at the upper-end of the range shown in Table~\ref{tab:pose_ranges}. We also decided to only consider the translational (normal and shear) components of force to simplify the control and analysis. Due to the extreme underactuation of the SoftHand, grasp stability is controlled through monitoring the distribution of forces across the grasp, as individual fingertips cannot be controlled. \textcolor{black}{Therefore, small changes in torsional shear make little difference in the situations considered in this paper where the object's centre of mass lies inside the grasp envelope; furthermore, their inclusion would greatly complicate the control strategy. Other manipulation scenarios where torsional shear may be appreciable would require extension of the control strategy.}  



For each tactile image sample, the sensor was brought into contact with the surface at the defined orientation and depth, then sheared across the surface to the defined shear displacement. The resulting tactile image was captured and the forces sampled from the FT sensor, with image and force capture processes running in parallel and synchronized by timestamp to ensure the correct data was extracted. Data captured from the FT sensor at each time step was taken as a moving average across 50 samples, then a Butterworth filter was used to further smooth the signal across the entire tap-and-shear movement for that point of data collection. This type of hybrid filtering approach is common when balancing local noise reduction and overall signal smoothing, and found to be important in this case due to a small signal-to-noise ratio. In these cases, the moving average filter smooths local high frequency noise first allowing the subsequent Butterworth filter to work more efficiently on the broader frequency components of the signal~\cite{shi2012application, kurapa2020hybrid}.

Overall, 3000 tactile images were taken for each sensor and were split 80/20 to give 2400 training images and 600 validation images per sensor. A separate test dataset of 600 images was also collected for each sensor. The test data is distinct from the validation data in that it is unseen by the model during training to provide an objective performance evaluation. The test set verifies that any apparent accuracy in the trained models will generalize to new, previously unseen data, ruling out overfitting.

\subsection{Contact Pose/Force Model for Multiple Sensors} \label{sec:init_training}
Initially, a model was trained using the data for each sensor with the convolutional neural network architecture from \cite{lepora2020optimal}. The training and model hyperparameters are given in Table~\ref{tab:hyperparams} (Appendix \ref{appA}) and are from previous work using a TacTip to estimate surface pose~\cite{lepora2020optimal}. Each CNN takes a $240\times135$\,px tactile image as input and outputs a $6$-component vector of pose and force predictions, namely $[z,\alpha,\beta,F_x,F_y,F_z]$ (as shown in Fig.~\ref{fig:robo_platform}c). As force should increase monotonically with translations in the $x$-, $y$- and $z$-directions, we expect that the model can use same tactile image features for force as for the pose counterpart from Ref.~\cite{lepora2020optimal}. Details of the network structure and hardware are given in Appendix~\ref{appA}.

\begin{figure}[t!]
        \centering
        \includegraphics[width = 0.48\textwidth]{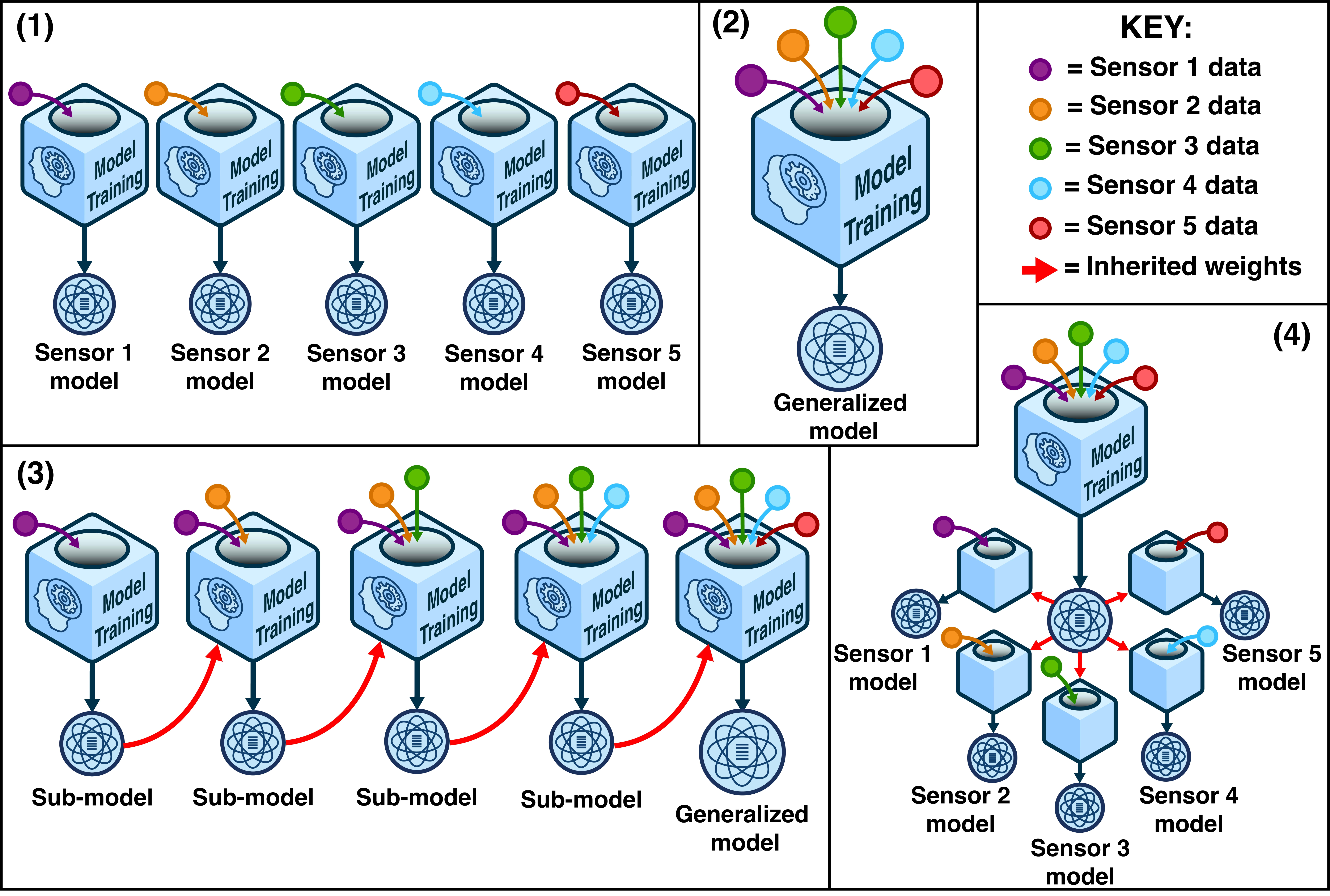}
        \caption{Different approaches for supervised learning of tactile pose and force prediction models based on convolutional neural networks. 1)~\textbf{Individual models}: 5 single models are trained on data from each sensor individually. 2)~\textbf{Aggregate model}: one model is trained on data aggregated from all 5 sensors. 3)~\textbf{Progressive transfer model}: one model is trained progressively on data from each of the 5 sensors in turn. 4)~\textbf{Standard transfer models}: 5 individual sensor models are trained on data from each sensor individually, from a pre-trained model trained on data from all 5 sensors.}
        \label{fig:transfer_training}
        \vspace{-1.5em}
\end{figure}

Since data from multiple sensors was available, we investigated whether transfer learning could be used to improve individual model performance. Initial investigations showed that a force prediction model for one sensor cannot be directly used to make accurate predictions for another, which we attribute to small physical discrepancies between them resultant of manufacturing and assembly. However, as the marker layout is similar for all sensors, we expect that some trained features would be common to all sensors regardless. Thus, we expect that a model that has inherited the weights and biases of a pre-trained model designed to work in the same feature space but from different sensors would be better at generalizing between sensors to improve overall accuracy and robustness.

Four learning approaches were compared (summarized in Fig. \ref{fig:transfer_training}). The baseline is an individual model approach; {\em i.e.} the same training method described above. Next, an aggregate approach was taken, which trains a model on a dataset from all sensors. Two typical transfer learning approaches were taken: firstly, progressive transfer learning (which trains an individual model successively with data from all sensors, inheriting weights and biases from the previous iteration~\cite{yu2022progressive}); and, secondly, standard transfer learning, which trains individual models from data for each sensor, inheriting the weights and biases of the aggregate model~\cite{zhuang2020comprehensive}. Once learned, instances of models may be sent to individual Jetson Nanos in the processing array and deployed in parallel.

\subsection{Control Framework} \label{sec:grasp_controller}
The controllers used in this study all use force feedback from the tactile sensors to affect a system response, whether that is grasp adjustment or a change in robot arm velocity. In this section, we will first introduce our grasp controller framework used in Experiments 1 and 2, followed by the velocity controller used in Experiment 3.

\begin{figure}[t!]
        \centering
        \includegraphics[width = 0.475\textwidth]{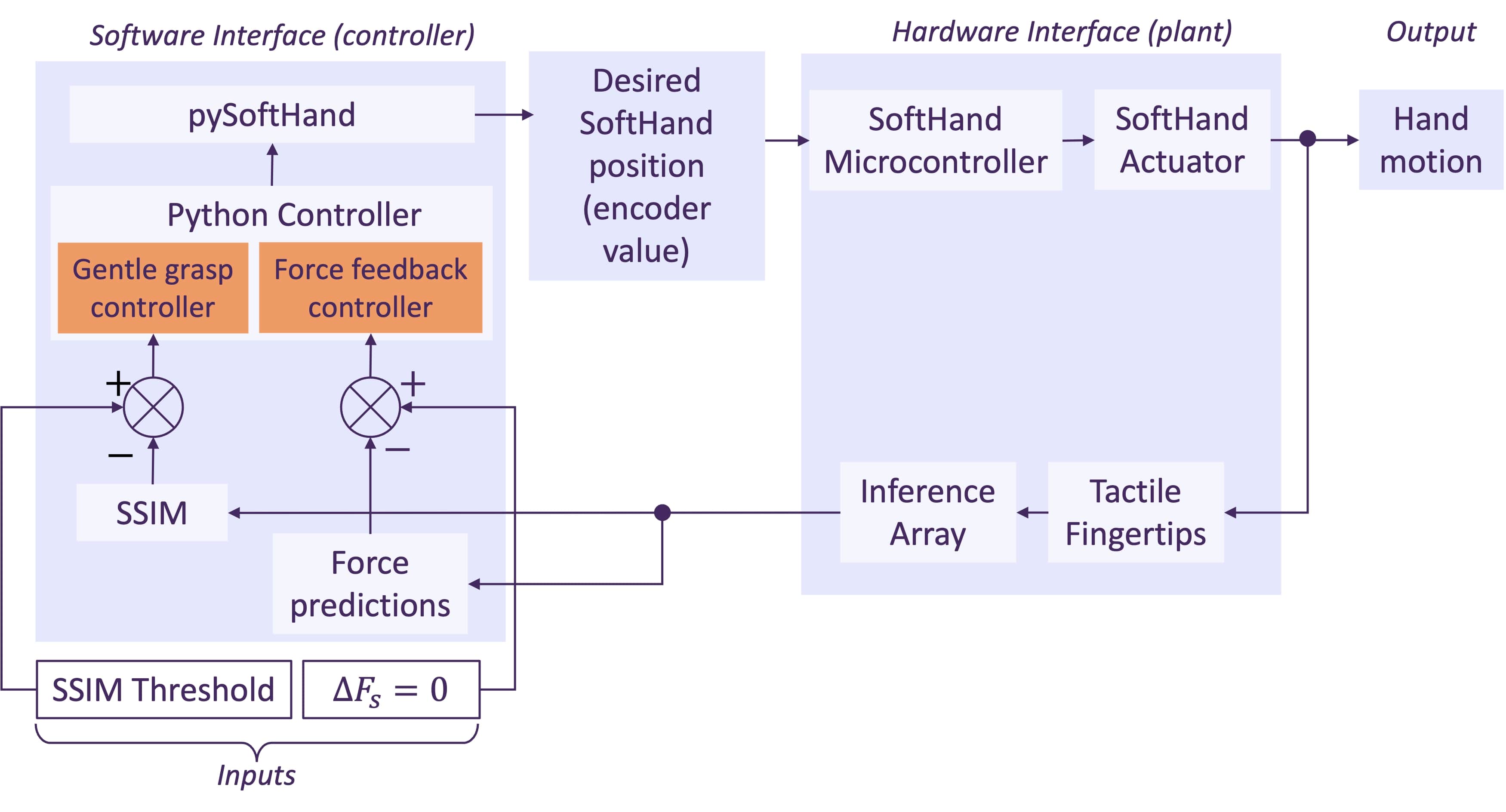}
        \caption{Shear-based grasp controller architecture. The gentle grasp controller uses an SSIM measure of contact deformation to establish a stable yet gentle grasp on an object. Meanwhile a force-feedback controller uses force predictions from the tactile sensors to modulate the grasp in response to external disturbances. These controllers feed into the hardware interface (plant) comprising the SoftHand microcontroller and tactile model prediction array that are depicted in more detail in Fig.~\ref{fig:robo_platform}.}
        \label{fig:control_diagram}
\end{figure}

The grasp controller used in this study aims to gently grasp an object and retain it in response to external disturbances by modulating the grasp force. The general architecture is shown in Fig. \ref{fig:control_diagram} and has two main components described below.

\textbf{Gentle grasp controller}: The objective of this controller is to establish and maintain a light contact, applying just enough force to retain it. This controller uses the average Structural Similarity Index Measurement (SSIM) \cite{wang2004image}, an established measure for contact detection in optical tactile
sensors \cite{james2021tactile,10161036} that functions by calculating a percentage similarity between two images. It is calculated by
\begin{equation}
	    {\rm SSIM}\left( \mathrm{Img},\mathrm{Img}_{0} \right)=\frac{\left( 2\mu_x\mu_y+c_1 \right)\left( 2\sigma_{xy}+c_2 \right)}{(\mathrm{\mu}_{x}^{2}+\mathrm{\mu}_{y}^{2}+c_1)(\mathrm{\sigma}_{x}^{2}+\mathrm{\sigma}_{y}^{2}+c_2),}
	    \label{eq:ssim}
\end{equation}
where $x$ and $y$ represent an $n \times n$ kernel of pixels applied as a sliding window over each image. $\sigma$ and $\mu$ represent the mean and covariance of each kernel calculation, with $c_1$ and $c_2$ acting as regularising constants to stabilize the division \cite{james2021tactile}. The SSIM is given as an averaged final value, SSIM$\ \in [0,1]$, where SSIM$\ =1$ indicates the images are identical and SSIM$\ =0$ indicates they have zero similarity~\cite{wang2004image}.

In the gentle grasp controller, the SSIM of a baseline, undeformed sensor image and the current sensor image is taken across all the sensors to establish and maintain a light contact using two sequential proportional controllers, each optimised for states where contact is and is not detected~\cite{10161036}. 

\textbf{Force feedback controller}: This controller uses force predictions from the tactile sensors to modulate grasp force after an initial grasp is established by the gentle grasp controller. \textcolor{black}{The change in shear forces, $\Delta F_x$ and $\Delta F_y$, are used as feedback variables, computed over one time step (17\,ms average). By defining a controller setpoint of $\Delta F_x=\Delta F_y=0$\,N and using the error from the setpoint as feedback, we describe control behavior that will minimize the change in shear force by modulating the grasp force, retaining the object in the grasp}. This is achieved by a standard PID controller. The tuning criteria for the gains of the force-feedback controller were to give a response which was sensitive to both fast and slow disturbances and would not drop nor damage the held object. Due to the highly underactuated nature of the SoftHand and the non-linearity of the dislocatable finger joints, the system cannot be easily assessed using traditional control engineering methods such as frequency response. \textcolor{black}{Consequently, the system was manually tuned to a point where the observed behavior met the quantitative criteria outlined in Section~\ref{sec:exp1_rev}}.

\begin{figure}[t!]
    \centering
    \includegraphics[width = 0.475\textwidth]{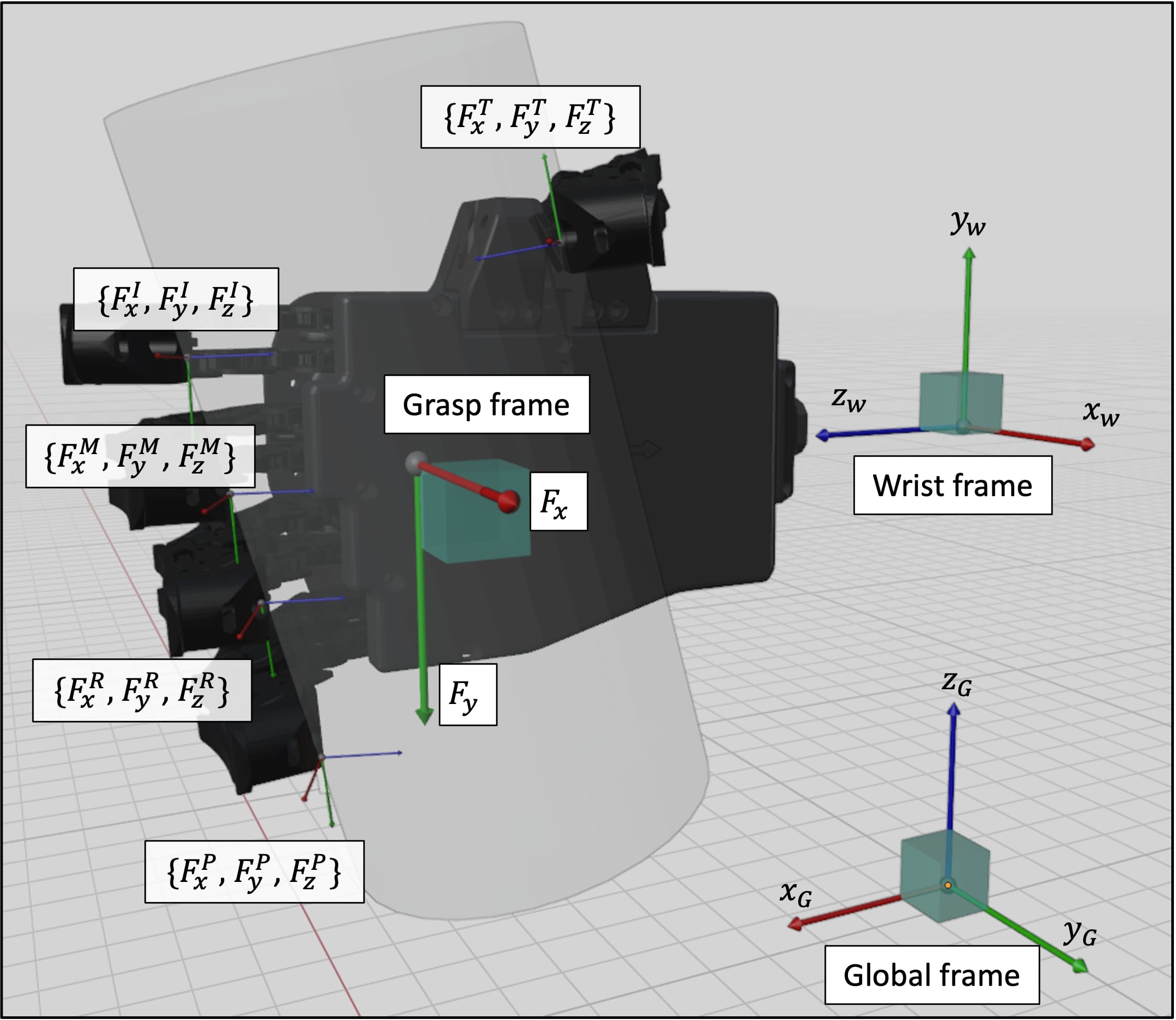} 
    \caption{Various frames of reference and force components used for controlling the Pisa/IIT SoftHand on the robot arm. The global frame $\{x_G,y_G,z_G\}$ is the base frame of the UR5 robot arm, with the hand in the wrist frame $\{x_w,y_w,z_w\}$ of that robot. The force predictions from the individual tactile sensors are in the frames of those sensors ({shown here as $\{x,y,z\}$ frames at each fingertip, where the superscript indicies P, R, M, I and T refer to the Pinky, Ring, Middle, Index and Thumb digits respectively}), which are consolidated into a single force vector by averaging the shear forces at each fingertip. The $x$ and $y$ axes for the force vector are assumed to align with the $x_w$ and $y_w$ axes of the wrist frame, allowing the orientation of the shear force vector to be resolved from the robot kinematics.}
\label{fig:graspframe}
\end{figure}

The underactuation and soft nature of the SoftHand also means the exact pose of the sensors relative to the global frame cannot be known. Therefore, we define a `grasp frame', defined as the average of all contact shear forces from all sensors, taking the form of a pair of resultant shear force vectors, $F_x$ and $F_y$, which align with the $x_G$ and $z_G$ axes of the global frame (as shown in Fig. \ref{fig:graspframe}) when the robot arm is in starting pose with the thumb facing upward and the palm surface parallel with the global $xz$-plane (see Fig.~\ref{fig:robo_platform}(a)). 

\subsubsection{\textbf{Static robot with external disturbance}} \label{control1}

The grasp frame has the same orientation with respect to the global frame as the wrist frame of the UR5 robot arm. Thus, the pose of the grasp frame relative to the global frame may be resolved as a translation from the wrist frame pose, which is known from the kinematics of the robot arm.

\textcolor{black}{To define the grasp frame, we use the sum of the shear forces across all fingers in both the $x$- and $y$-directions}. The rate of shear force at each time step, $\Delta F_x$ and $\Delta F_y$, is found by calculating the backward finite difference of the shear force in each direction. This is taken across an interval of 50 samples to reduce correlated noise in the signal.

\subsubsection{\textbf{Non-Static robot with external disturbance}} \label{control2}

When the robot arm is in motion, the dominant shear force direction will change with the orientation of the grasp relative to gravity. This presents several unique challenges compared to the previous static case, with the aim being to continually maintain a stable yet gentle grasp regardless of the orientation with respect to gravity or the variability of an object's mass. 

Unlike the static case, the shear forces in the $x$-direction must be considered along with those in the $y$-direction, as the $F_x$ vector no longer remains perpendicular to gravity at all times. This complexity is handled by adding an additional controller with the same PID structure that was applied to the $y$-shear in the static case with different gains to control shear in the $x$-direction. It was found to be necessary to have separate controllers handling the $x$- and $y$-shear (as opposed to one controller considering the overall tilting angle) as the grasping behavior is highly dependent on the dominant shear axis. An aggregate controller was initially trialled, however was found to be more difficult to tune to the desired behavior than two independent controllers.

Because the Pisa/IIT SoftHand is underactuated, a change in grasp position tends to induce a small shear force in the non-dominant shear direction due to the fingertips moving on the contact surface. Hence, using the same control architecture as described above can cause the hand to grasp with excessive force, because it will interpret this additional shear force as a result of the grasp disturbance rather than a side-effect of grasp force modulation. To address this problem, we scale each shear force vector according to its orientation relative to gravity, which is crucial as it is necessary to know when the hand is upside-down. Therefore, we use the following as inputs to each controller:
\begin{equation}
    \label{eqn:ux}
    u_x = S_x\Delta F_x, \ \
    u_y = S_y\Delta F_y,
\end{equation}
where $S_x$ and $S_y$ are scaling factors
\begin{equation}
\label{eqn:sx}
    S_x = 1 - \frac{2\theta_x}{\pi}, \ \
    S_y = 1 - \frac{2\theta_y}{\pi}, 
\end{equation}
and $\theta_x$ and $\theta_y$ are the angles between gravity and $F_x$ and $F_y$ respectively (see derivation in Appendix \ref{appB}). 

To find $\theta_x$ and $\theta_y$ given a static XYZ (SXYZ) Euler rotation $(\alpha, \beta, \gamma)$ (taken as the pose of the grasp frame relative to the robot's base frame), we first find the product of the rotation matrices for each axis,
\begin{equation}
    \rm{R}_{SXYZ} = \rm{R}_z(\gamma) \ \rm{R}_y(\beta) \ \rm{R}_x(\alpha).
\end{equation}
We can then find the rotated $x$ and $y$ axes,
\begin{equation*}
    \rm{x}' = \rm{R}_{SXYZ} \cdot \begin{bmatrix} 1 & 0 & 0 \end{bmatrix}^T, \ 
    \rm{y}' = \rm{R}_{SXYZ} \cdot \begin{bmatrix} 0 & -1 & 0 \end{bmatrix}^T.
\end{equation*}
If the gravity vector, ${\mathbf{g}} = \begin{bmatrix} 0 & 0 & -1 \end{bmatrix}^T$, is in the $-z_G$ direction of the global frame (as shown in Fig. \ref{fig:robo_platform}), then $\theta_x$ and $\theta_y$ can be found using
\begin{equation} \label{eqn:thetas}
    \!\!\!\theta_x = \arccos\left(\frac{{\bold{g}} \cdot {\bold{x}'}}{\|{\bold{g}}\| \  \|{\bold{x}'}\|}\right),
    \theta_y = \arccos\left(\frac{{\bold{g}} \cdot {\bold{y}'}}{\|{\bold{g}}\| \  \|{\bold{y}'}\|}\right).
\end{equation}

\subsubsection{\textbf{Control of robot under externally-applied forces}} \label{control3}

As discussed in Section \ref{sec:background}, more traditional methods of robotic control through force sensing (such as a FT sensor in the wrist of the robot) do not apply well to the SoftHand due to the elasticity of the finger joints. By using tactile feedback, we are able to translate disturbances to the grasp into a velocity control signal for the robot, instructing it to move with the applied force. Assuming the SoftHand is grasping an object, an external force can then be applied to the object. The resultant magnitude and direction of this applied force relative to the global frame can be resolved and translated into a velocity control signal for the UR5 robot arm, whereby it will move in the direction of the applied force at a velocity proportional to the magnitude of the applied force.

Here we consider transforming the 3D force vector $(F_x, F_y, F_z)$ into a linear velocity vector $(\dot{x},\dot{y},\dot{z})$. Only linear velocities are considered, as only linear forces, not torques, are used in the prediction process, constraining the robot to only translational movements. The components of the velocity vector align with the $x_{\rm w}$, $y_{\rm w}$ and $z_{\rm w}$ axes of the wrist frame of the robot (Fig. \ref{fig:graspframe}). As previously mentioned, the fingertip positions and orientations with respect to the global frame cannot be known due to the lack of proprioception in the finger joints; thus, the force components $F_x$ and $F_y$ are taken as the sum from each fingertip. \textcolor{black}{The force component $F_z$ is defined as the sum of the normal force experienced by all fingers and is assumed to act perpendicularly to $F_x$ and $F_y$ (Fig. \ref{fig:graspframe}).}

The velocity control of the robot arm is then given by a system of linear equations that produces a linear velocity vector acting in the the resultant direction of the applied forces:
\begin{equation}
\begin{bmatrix}
\dot{x} \\
\dot{y} \\
\dot{z}
\end{bmatrix}
=
\begin{bmatrix}
        k_x(F_x) & 0 & 0 \\
        0 & k_y(F_y)  & 0 \\
        0 & 0 & k_z(F_z) 
    \end{bmatrix} \begin{bmatrix}
F_{x} \\
F_{y} \\
F_{z} 
\end{bmatrix},
\label{eqn:ex3}
\end{equation}
\textcolor{black}{where the controller input is the predicted force vector $(F_x,F_y,F_z)$ at the current time-step.} The velocity control is handled by the proprietary controller on the UR5 robot arm.

To scale the output proportionally to the magnitude of the applied force, the ratio of the applied force relative to its maximum value (Table \ref{tab:pose_ranges}) is multiplied by a base proportional gain $k_0$ to create a dynamic proportional gain $k(F)=k_0\lvert F\rvert/F_{\rm max}$, where $F_{x,{\rm max}} = F_{y,{\rm max}} = 20$, $F_{z,{\rm max}} = 60$.
The effect of this is that smaller signal perturbations, such as that from sensor noise, are diminished to prevent unwanted movement. \textcolor{black}{(Note that Eq. \ref{eqn:ex3} results in an $F|F|$ term in the controller, which as described above results in a force-modulated gain acting on a force.) Then the base proportional gains $k_{x,0}$, $k_{y,0}$ and $k_{z,0}$ were tuned such that a user input magnitude greater than $\pm0.2$\,N would initiate robot movement, so as to eliminate sudden movements at low input forces. This procedure results in a controller tuned to tasks with these force ranges.}

For each controller presented in this section, once the control parameters were tuned it was found that the desired controller behavior generalized to objects of differing geometry and stiffness. This is as to be expected given results from previous work on grasp controller development with this sensorimotor platform~\cite{10161036}.

\section{Experimental Methodology} \label{sec:exp_methods}

\begin{algorithm}[b!]
\caption{Static Grasp Stabilization}
\begin{algorithmic}[1]
    \State establishGrasp()\{
        \While{grasped = False} 
            \State SSIM $\gets$ sensors.getSimilarity()
            \State pos $\gets$ softhand.getPosition()
            \If{SSIM $\geq$ contactThreshold}
                \State softhand.setPosition(pController0(pos))
            \Else 
                \State softhand.setPosition(pController1(SSIM))
            \EndIf
            \If{SSIM = setPoint}
                \State grasped = True
            \EndIf
        \EndWhile \\
        \}
    \State $i = 0$
    \While{True}
        \State $F_y \gets$ extractShear(predictions)
        \If{$i \geq \delta$}
            \State $\Delta F_y \gets F_y[i] - F_y[i-\delta]$
            \State softhand.setPosition(pidController($\Delta F_y$))
        \Else
            \State doNothing()
        \EndIf
        \State $i++$
    \EndWhile
\end{algorithmic}
\label{al:static}
\end{algorithm}

\subsection{Experiment 1: Static robot with external disturbance} \label{sec:exp1_rev}
For the first experiment, we mount the tactile SoftHand on a UR5 robot arm, and seek to maintain a grasp whilst grasping a cup as rice is added to the cup to disturb that grasp. A deformable paper cup is used to exhibit how the system is able to grasp an object with just enough force to maintain stability without crushing, which can only be reliably achieved with such a hand via tactile sensing and control~\cite{10161036}. Initial testing found that the paper cups used in our experiments permanently deformed (i.e were crushed) under $F_z$ loads of approximately 9\,N. Additionally, under initial gentle grasp conditions the cup began to slip after approximately 20\,g of rice had been added.  We use the controller that retains a static grasp under an external disturbance (Section~\ref{control1}). For this experiment, the change in shear force $\Delta F_x$ is assumed to be 0 as the hand is grasping the cup whilst keeping a vertical orientation; therefore, the $x$-component of the force (as seen in Fig. \ref{fig:graspframe}) is perpendicular to gravity and has no effect on the system.

In the procedure for this experiment (see Algorithm \ref{al:static}), the hand first closes onto the cup and maintains a stable, gentle grasp, then the control switches to the force feedback controller. At this point, rice is added to the cup to destabilize the grasp. Three different masses of rice were used (100\,g, 200\,g, 300\,g) to test that the controller responds as intended to a range of disturbance conditions. \textcolor{black}{An initial test with an empty cup and constant grasping force of $F_z \approx 2$\,N was conducted as a baseline, which failed the task by deforming the cup. Higher grasp forces in the task are only possible when there is mass in the cup due to the outward pressure exerted, which is why a dynamic control strategy is necessary.} Evidence of this is shown in the supplementary media.

\textcolor{black}{To ensure that the controller gives a desirable response to both fast and slow disturbances, step and ramp input tests were performed. The tuning criteria for these tests are as follows:\\
\noindent\textbf{Step input}: The error signal ($\Delta F_y$) should settle to a setpoint (0$\pm$0.05\,N/s) within 0.5\,s for the maximum input (300\,g rice). The plant response (SoftHand's motor position) will exhibit some oscillation due to elasticity in the system; however, the grasp force ($F_z$) should not exceed the cup's crush force (9\,N).\\
\noindent\textbf{Ramp input}: The error signal should not exceed 0.25\,N/s. The plant response should increase steadily throughout the test with a steeper gradient for higher masses of rice. The grasp force should track the plant response, yet not exceed the crush force.\\
For both cases, the cup may slip in the grasp, but must not be dropped. The tuned gains are given in Table \ref{tab:control_params}, Appendix \ref{appC}.}

\subsection{Experiment 2: Moving robot with external disturbance} \label{sec:exp_2}
For the second experiment, we examined the situation where the orientation of the hand is not static while the grasp is destabilized. The rice is poured into the cup like in experiment 1, then the arm is rotated about the grasp frame such that the rice is poured out of the cup. During this movement, the centre of mass of the cup will shift due to the fluid motion of the rice, before the mass of the entire object (rice plus cup) changes as the rice is poured from the cup. We use the controller for a non-static robot with external disturbance (Section~\ref{control2}) \textcolor{black}{with gains described in Table \ref{tab:control_params}, Appendix \ref{appC}}.

In the procedure for this experiment (see Algorithm~\ref{al:dynamic}), the robot is rotated about the grasp frame (Fig. \ref{fig:graspframe}) by an angle that is sufficient to pour the rice out of the cup. There are a family of possible rotations $(\alpha, \beta, \gamma)$ that could be applied to the hand. However, a pure $\gamma$ rotation around the $z$-axis is actually the most challenging, as the friction of the grasp is opposing the entire mass of the cup for the whole motion; {\em i.e.} there is no point where the mass is partially or fully supported by either the thumb or fingers. A pure $\gamma$ rotation also exhibits instances where the angles between the $x$ and $y$ shear force vectors and gravity, $\theta_x$ and $\theta_y$ (Eq.~\ref{eqn:thetas}) are the dominant shear forces in the interaction. For these reasons, we consider a rotation $\gamma=120^\circ$ to pour the rice from the cup.

\begin{algorithm}[t!]
\caption{Dynamic Grasp Stabilization}
\begin{algorithmic}[1]
    \State establishGrasp()
    \State $i = 0$
    \While{True}
        \State pose $\gets$ robot.getPose()
        \State $\alpha, \beta, \gamma\gets$ extractRotation(pose)
        \State $\theta_x, \theta_y \gets$ eulerRotation($\alpha, \beta, \gamma$)
        \State $F_x, F_y \gets$ extractShear(predictions)
        \If{$i \geq \delta$}
            \State $\Delta F_x \gets F_x[i] - F_x[i-\delta]$
            \State $\Delta F_y \gets F_y[i] - F_y[i-\delta]$
            \State $u_x \gets \Delta F_x S_x$
            \State $u_y \gets \Delta F_y S_y$
            \State $u \gets$ pidControllerX($u_x$) $+$ pidControllerY($u_y$)
            \State softhand.setPosition($u$)
        \EndIf
        \State $i++$
    \EndWhile
\end{algorithmic}
\label{al:dynamic}
\end{algorithm}

\begin{figure*}[b!]
        \centering
        \includegraphics[width = \textwidth]{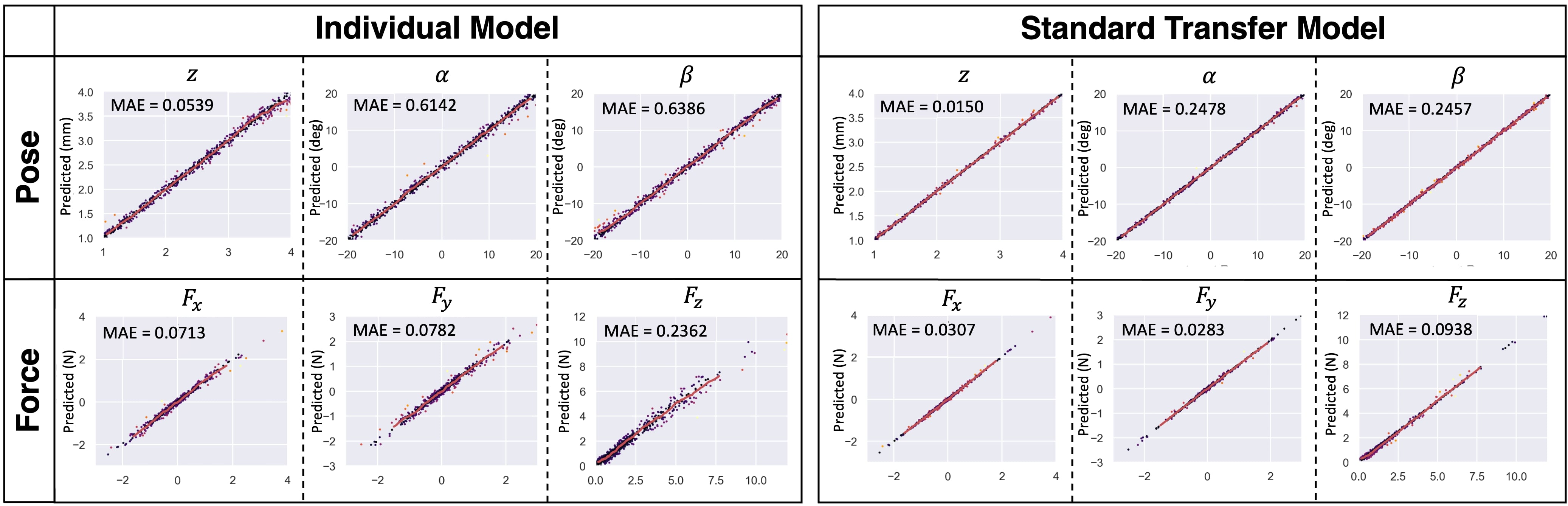}
        \caption{Tactile pose and force model predictions. The individual scatter plots show the predicted pose and force plotted against the ground-truth labels taken upon data collection. Also shown is the Mean Absolute Error (MAE) of prediction accuracy. The same 5 test datasets for each fingertip are used in both cases, with both the individual models and the standard transfer models applying the individually trained model for each fingertip to the matching test dataset.}
        \label{fig:exp1_result}
\end{figure*}

\subsection{Experiment 3: Tactile-Driven Leader-Follower Task}
For the final experiment, we consider a tactile-driven leader-follower task where tactile information is translated into motion of the UR5 robot arm's end-effector in the global frame (see Fig. \ref{fig:graspframe}). The aim of this experiment is to demonstrate the level of sensitivity and control that may be achieved through tactile sensing with an underactuated hand on a robot arm. The robot operates under the controller that moves the end-effector under externally-guided forces (Section~\ref{control3}) \textcolor{black}{using the gains described in Table \ref{tab:control_params}, Appendix \ref{appC}}. The aim of this task is for the operator to be able to input exerted forces that will result in a smooth, stable velocity response of the hand on the arm; i.e. no drifting when stationary and the robot should come to a stop when the force is removed.

\textcolor{black}{The experimental setup consisted of the SoftHand mounted on the UR5 robot arm, grasping a 3D-printed square profile stimulus of constant cross section (40\,mm $\times$ 40\,mm $\times$ 170\,mm). This object is split, with an ATI Mini40 force-torque sensor in the center to create a sensorized object from which to measure ground-truth force}. A human operator applies a force on the object, detected by the tactile fingertips which under controller \ref{control3} act as an input to the velocity controller. The operator applies forces to a held object such that the robot end-effector follows a trajectory along each axis individually. The instructions given to the operator were to apply a force to the object such that the robot end-effector traces a path that passes through setpoints keeping the motion as close to the respective axis of motion as possible, in the order of setpoints: origin$\rightarrow\!\!x$-axis$\rightarrow$origin$\rightarrow\!\!y$-axis$\rightarrow$origin$\rightarrow\!\!z$-axis$\rightarrow$origin.

\section{Results}
In this section, we examine the results of the experiments. Firstly, we investigate the performance of the pose and force prediction models discussed in Section \ref{sec:data_collect}, and how performance may be improved by transfer learning. Next, we evaluate the performance of the force-sensitive grasp controller presented in Section \ref{sec:grasp_controller} through the experiments presented in Section \ref{sec:exp_methods}: first, the system's ability to grasp and retain a paper cup in response to different degrees of external disturbance; second, a pouring task; and finally, a leader-follower task driven entirely by tactile feedback. These use cases are synergistic as they all utilize shear-force feedback from the tactile sensors to affect a manipulation task involving a human operator. As such, the applications may be combined to give a complete system, capable of both grasp adjustment in response to external disturbance and manual repositioning from external force input.

\subsection{Contact Pose and Force Prediction Results}

\begin{table*}[t!]\centering
\caption{Average MAEs over all five microTac test sets (individual model vs multi-model approaches - see Fig.~\ref{fig:transfer_training}). The highest performance improvements over the baseline (individual model) are highlighted in bold.}
\begin{tabular}{|c|cccc|}
\hline
\multirow{2}{*}{\textbf{Prediction parameter}} & \multicolumn{4}{c|}{\textbf{Learning Methods - Average MAE (\% Improvement on baseline)}}                                                                                                     \\ \cline{2-5} 
                                               & \multicolumn{1}{c|}{\textbf{Individual Model (baseline)}} & \multicolumn{1}{c|}{\textbf{Aggregate Model}} & \multicolumn{1}{c|}{\textbf{Progressive Transfer}} & \textbf{Standard Transfer} \\ \hline
{$z$ (mm)}                              & \multicolumn{1}{c|}{0.051}                                & \multicolumn{1}{c|}{0.019 (63\%)}             & \multicolumn{1}{c|}{0.026 (49\%)}                  & \textbf{0.014 (72\%)}      \\
{$\alpha$ (deg)}                        & \multicolumn{1}{c|}{0.543}                                & \multicolumn{1}{c|}{0.257 (53\%)}             & \multicolumn{1}{c|}{0.291 (46\%)}                  & \textbf{0.245 (55\%)}      \\
{$\beta$ (deg)}                         & \multicolumn{1}{c|}{0.606}                                & \multicolumn{1}{c|}{0.263 (57\%)}             & \multicolumn{1}{c|}{0.308 (49\%)}                  & \textbf{0.245 (60\%)}      \\
{$F_x$ (N)}                             & \multicolumn{1}{c|}{0.075}                                & \multicolumn{1}{c|}{0.037 (51\%)}             & \multicolumn{1}{c|}{0.045 (40\%)}                  & \textbf{0.032 (57\%)}      \\
{$F_y$ (N)}                             & \multicolumn{1}{c|}{0.078}                                & \multicolumn{1}{c|}{0.038 (51\%)}             & \multicolumn{1}{c|}{0.047 (40\%)}                  & \textbf{0.031 (60\%)}      \\
{$F_z$ (N)}                             & \multicolumn{1}{c|}{0.222}                                & \multicolumn{1}{c|}{0.096 (57\%)}             & \multicolumn{1}{c|}{0.126 (43\%)}                  & \textbf{0.080 (64\%)}      \\ \hline
\end{tabular}
\label{tab:learning_improvement}
\end{table*}

\begin{table}[b!]
\centering
\caption{A comparison of the training time (in hours) for each learning method considered}
\arrayrulecolor{black} 
\begin{tabular}{|>{\centering\arraybackslash}m{3cm}|>{\centering\arraybackslash}m{3cm}|} 
\hline
\textbf{Training Method} & \textbf{Average Training Time (hours)}  \\ 
\hline
Individual Model         & 0.51 (2.55 total)                                   \\ 
Aggregate Model          & 0.73                                    \\ 
Progressive Transfer     & 4.22                                    \\ 
Standard Transfer        & 1.31 (3.28 total)                                   \\ 
\hline
\end{tabular}
\label{tab:training_time}
\end{table}

\begin{figure}[b!]
    \centering
    \includegraphics[width = 0.45\textwidth]{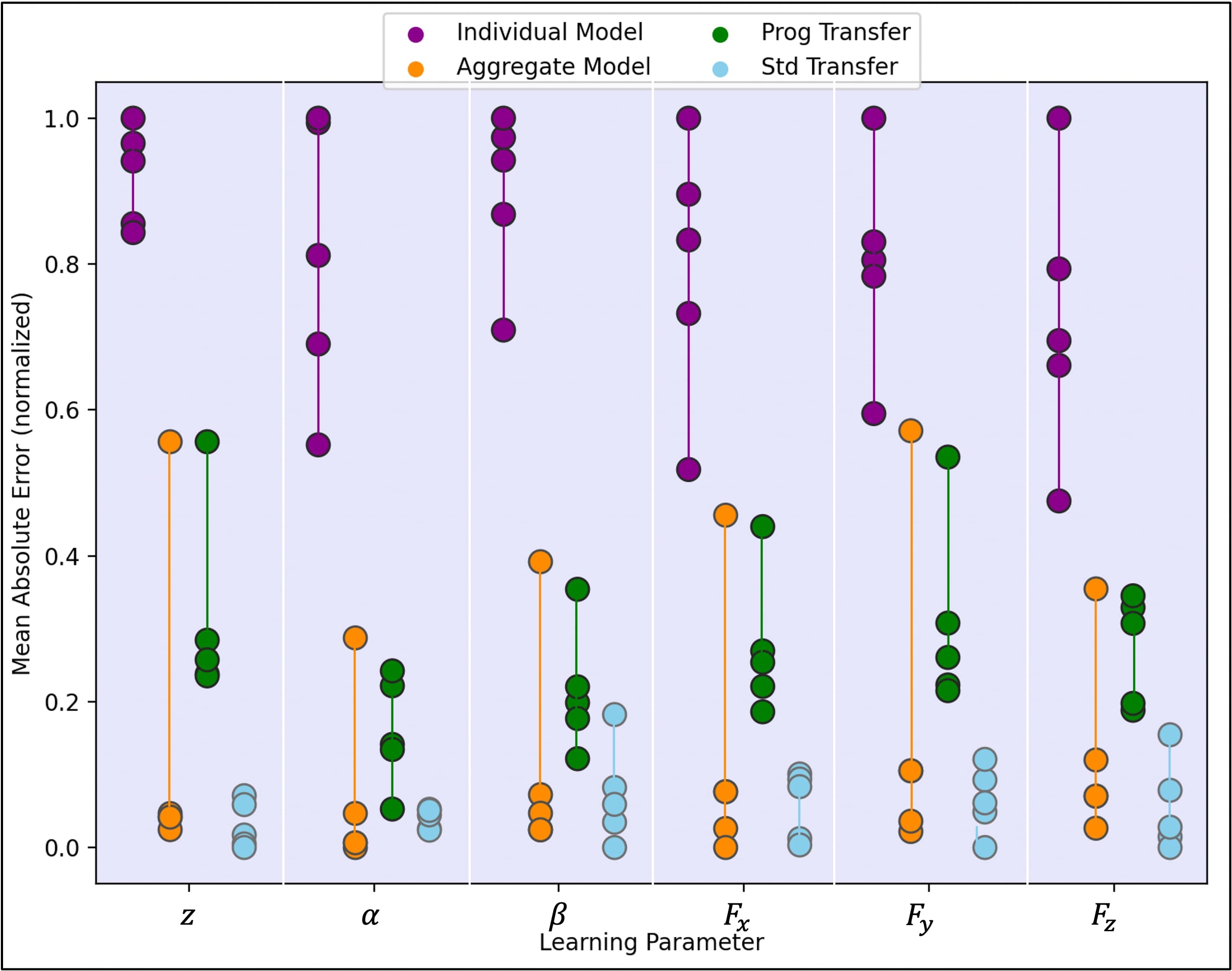}
    \caption{Comparison of the model prediction accuracies for pose and force. Four training approaches were considered (see Fig.~\ref{fig:transfer_training}), and the model accuracies shown for each of the six pose and force predicted parameters. For each model and parameter, five mean absolute errors (MAEs) are shown corresponding to test data from each of the five fingertips. Only the standard transfer method has a low MAE for every fingertip.}
    \label{fig:compare_method}
\end{figure}
In this experiment, we compare the performance of a pose and force estimation model using data taken from a single sensor against the transfer learning methods presented in Section \ref{sec:data_collect}.

To establish a baseline, we first used the individual-model approach depicted in Fig.~\ref{fig:transfer_training} to predict contact pose and force components from a test data set. The test set was sampled from the same sensor used to collect the training and validation data. For pose component prediction, the performance matches what we would expect for a TacTip in this range \cite{lepora2020optimal} (Fig.~\ref{fig:exp1_result}). The force predictions are non-uniformly distributed, as expected due to the samples which will have encountered a slip event during data collection. However, because the data is labelled post-slip, this effect manifests itself simply as this non-uniformity as opposed to resulting in noisy predictions (as was the effect in previous work on predicting pose and shear~\cite{lloyd2023pose}). Hence, this result shows that we are able to predict contact pose to a similar accuracy as in previous work with this method as well as contact shear force without needing to reduce the noise in the predictions (which required a model of the state dynamics~\cite{lloyd2023pose}).

Next, we evaluate the performance of the aggregate and transfer-learning methods. For a fair comparison, each method is tested on the same test data from the 5 fingertips. We found that these methods outperformed the individual-model method, all having lower mean absolute errors (MAEs) for each individual sensor (Fig. \ref{fig:compare_method}) and on average (Table~\ref{tab:learning_improvement}). Additionally, we note that whilst the aggregate model performs better overall than the individual-model method, it results in a large amount of variation in accuracy across the sensors. The outliers in each parameter for the Aggregate Model (Fig.~\ref{fig:compare_method}) are all from the same sensor, which highlights an issue with this method: it may favor accuracy for particular sensors rather than spreading that accuracy across all sensors. This issue does not happen with the transfer learning approaches, where performance is maintained at each training stage. 

Table~\ref{tab:training_time} shows the average training time for each method when run using CUDA on an Nvidia RTX 2060 GPU. As expected, the individual model has the fastest training time for a single sensor due to the relatively small dataset, but must be repeated five times for all sensors. The same is true with the standard transfer method, as this must also be repeated per sensor. The aggregate model is the fastest overall, as whilst it considers data from all sensors the training is parallelized to generate one generalized model applicable to all sensors. The progressive transfer method also results in a generalized model, however it takes significantly longer to train as the sensor data is introduced in sequential training stages.

Overall, the standard transfer-learning method yielded a consistent performance across the sensors with the lowest MAEs (Fig.~\ref{fig:compare_method}) and the lowest average MAEs (Table~\ref{tab:learning_improvement}). Therefore, this method is used in the grasping experiments. Quantitatively, the standard transfer learning model was able to improve the accuracy of predictions over 70\% compared to the individual-model approach (Table \ref{tab:learning_improvement}). Considering the prediction error plots for the standard transfer-learning method (Fig.~\ref{fig:exp1_result}), it is clear that this model gives much better predictions than the individual models. The reason for this is that as with any deep learning problem, more training data will generally result in higher accuracy~\cite{sun2017revisiting}. The benefit of training models on data from different sensors is that whilst accuracy will improve from a larger training dataset, the resultant models will become desensitized to non-systematic qualities of the feature space, resulting in better generalization~\cite{peng2019domain}.

\subsection{Experiment 1: Static Robot with External Disturbance}

In this experiment, we seek to evaluate the baseline performance of the grasp controller presented in Section \ref{sec:grasp_controller}. The desired control behavior is to grasp a paper cup and retain it without crushing in response to an external disturbance. To investigate this, the tactile SoftHand grasps the cup whilst mounted on the UR5 robot arm in the neutral pose (Fig.~\ref{fig:robo_platform}) and remains static whilst differing masses of rice are poured into the cup to disturb the grasp.

\begin{figure}[t!]
        \centering
        \includegraphics[width = 0.48\textwidth]{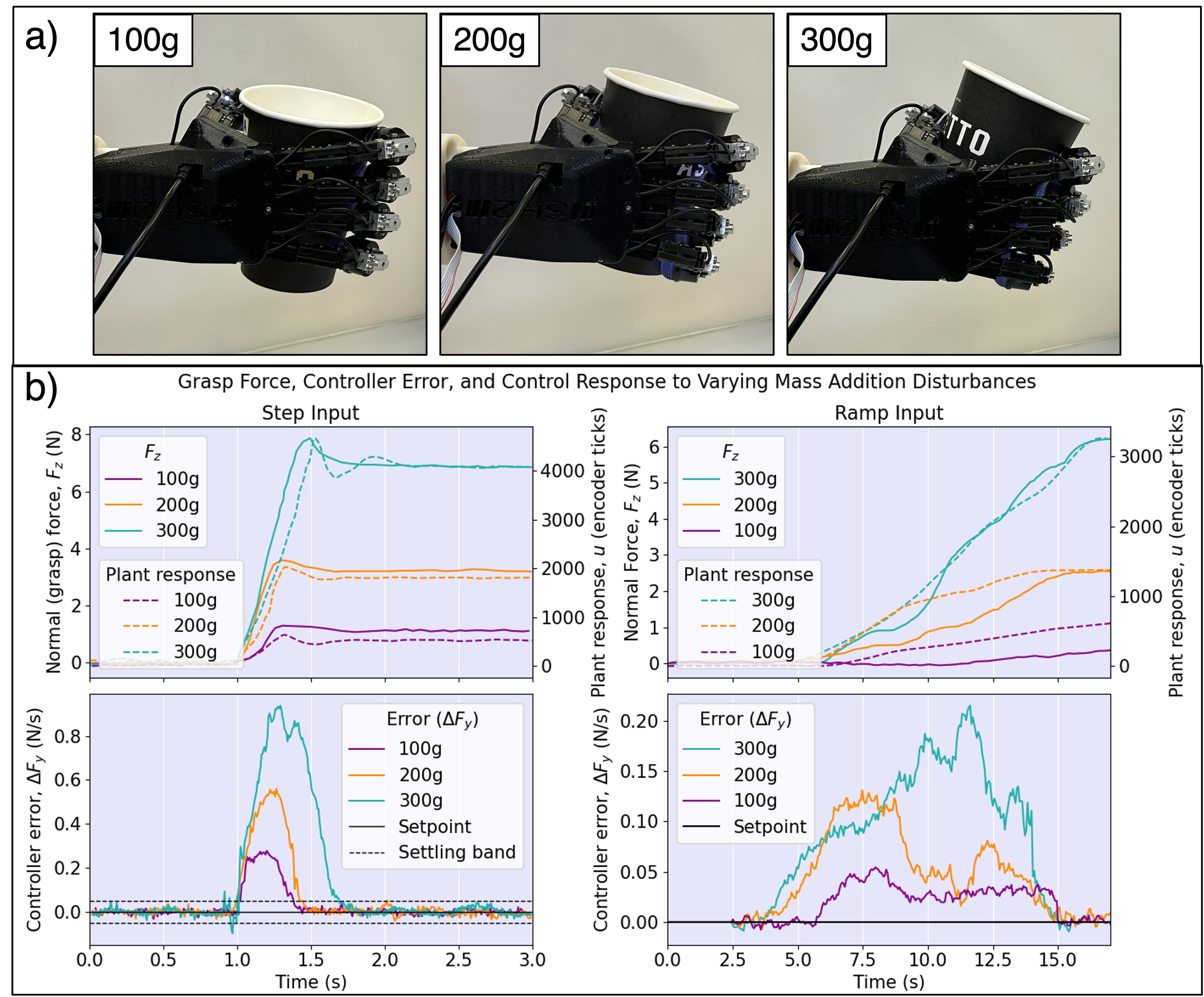}
        \caption{\textcolor{black}{Experiment 1 results: grasping a paper cup and retaining it without crushing in response to different masses of rice being poured into the cup. \textbf{(a)} Photos of the hand grasping the cup when the controller has achieved a set point after the rice being poured; note how the SoftHand exhibits greater finger displacement for higher masses. \textbf{(b, upper left)} normal (grasp) force, $F_z$ changing over time in response to step inputs of different magnitude. This plot also shows how the plant response, $u$ (actuator movement in the SoftHand) increases proportionally with $F_z$. Normal force is taken from predictions from the tactile sensors. \textbf{(b, lower left)} Controller error change over time. The error increases more for higher masses, indicative of the greater induced shear force. Once the input concludes, the error settles in an average time of 0.34\,s. Note how in each case the grasp stabilizes after a few seconds to a steady state with grasp force dependent on the weight of the poured rice. \textbf{(b, upper right)} Grasp force and plant response over time to ramping inputs. Once again, we see a more pronounced response to greater magnitudes of disturbance. \textbf{(b, lower right)} Controller error over time (ramp input). The controller's reaction to the disturbance keeps the error below the desired threshold of 0.25\,N/s.}}
        \label{fig:exp1_1_results}
\end{figure}

In total, three masses of poured rice (100\,g, 200\,g and 300\,g) up to the maximum capacity the cup could hold, were used to disturb the grasp. Five tests were performed for each mass of rice, and in each the cup was successfully retained in the grasp without slippage or being crushed. \textcolor{black}{Two sets of tests were performed to test the responses to step and ramp inputs. In each case, the settling time is the time to fall within the settling band defined between $\pm0.05$\,N/s (see Fig.~\ref{fig:exp1_1_results} b) and the steady-state error of the response is measured as the controller error, $\Delta F_y$, after settling (averaged over 2 seconds).}

\textcolor{black}{For the step input test, each mass of rice was added as quickly as possible to the cup (results in Fig.~\ref{fig:exp1_1_results}). The average settling time of 0.32\,s is well below the 0.5\,s threshold (described in Section \ref{sec:exp1_rev}) and the steady-state error is 0.0023\,N/s. The stability exhibited by the controller is also good, with minimal oscillatory behavior despite the compliant nature of the mechanism. The oscillations that occur were not considered problematic as they settle quickly and the grasp force does not exceed the 9\,N threshold.}

\textcolor{black}{For the ramp test, the mass is added gradually over a period of 7-9 seconds. The plant response and grasp force increase steadily in response to the disturbances (Fig.~\ref{fig:exp1_1_results}). The controller error also remains below the previously defined threshold of 0.25\,N/s, showing good error-tracking.}

\textcolor{black}{Note that the plant response can take longer to settle than the controller error, due to viscoelastic relaxation from the elastic nature of the cup, tactile skin and finger joints. This is acceptable as the quantity being controlled is $\Delta F_y$. Also, the hand does not exert a normal force greater than 9\,N (the baseline force that crushes the cup; see Section \ref{sec:exp1_rev}).} 

When observing the average normal force exerted on the cup, the controller behaved predictably for each disturbance case: exerting more force the greater the amount of added mass (Fig.~\ref{fig:exp1_1_results}). These results show that the grasp controller maintains an appropriate level of force and is able to handle delicate objects without damaging them, even in response to a disturbance. It can also be seen how the plant response, $u$, tracks the resulting grasp force, $F_z$.

We note how the adaptive synergy of the SoftHand interacts with the cup under the different loads (Fig. \ref{fig:exp1_1_results}a). The synergy of the SoftHand's grasping motion leads with the pinky finger, with each subsequent finger following in succession. When grasping the cup with more force, this causes the cup to pivot about the thumb, tilting away from the finger leading the grasp synergy (the pinky). This interaction helps stabilize the grasp, as some of the mass of the cup is transferred on to the upward-facing side of the leading fingers, partially supporting it against gravity. This behavior also illustrates another reason why the adaptive synergy of the hand is important, because it works in cooperation with the changing grasping force controlled by the tactile feedback from the fingertips.

\begin{figure*}[p]
        \centering
        \includegraphics[width = 0.85\textwidth]{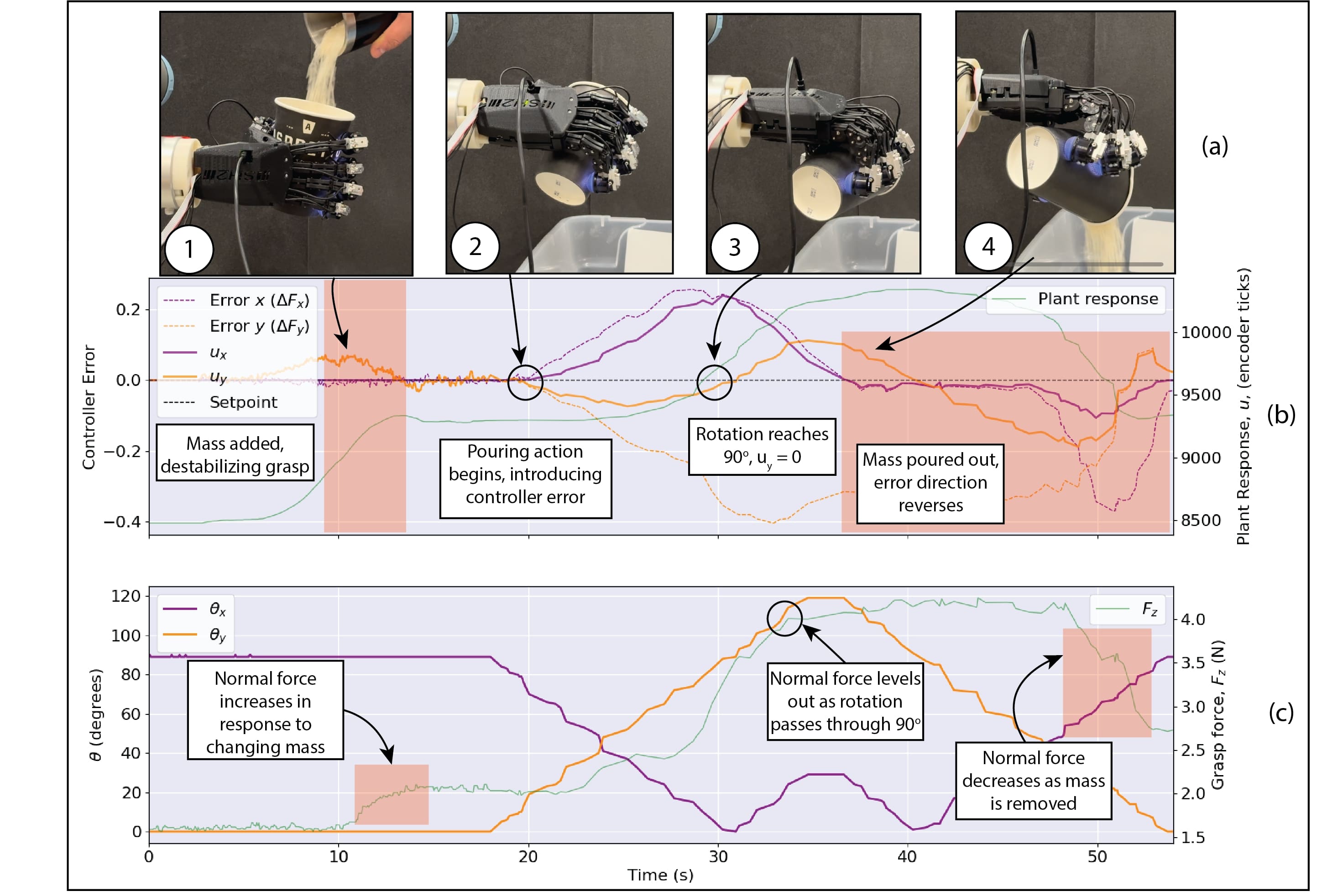}
        \vspace{-.5em}
        \caption{Experiment 2 results: mass is poured into a cup, which is then tilted to pour the rice out of the cup, with the hand adjusting its grasp according to the rate of change shear force. (a) Images of the hand during the pouring task. (b) The rate of change of shear force ($\Delta F_x$, $\Delta F_y$), equivalent to controller error, and the scaled values used as controller inputs ($u_x$, $u_y$). (c) Orientation angle change between ($\Delta F_x$, $\Delta F_y$) and gravity ($\theta_x$, $\theta_y$) throughout the task, and how the normal force changes over time, which serves to demonstrate how the grasp tightens and slackens during the task.}
        \label{fig:exp_2_result}
        \vspace{1em}
        \centering
        \includegraphics[width = 0.85\textwidth]{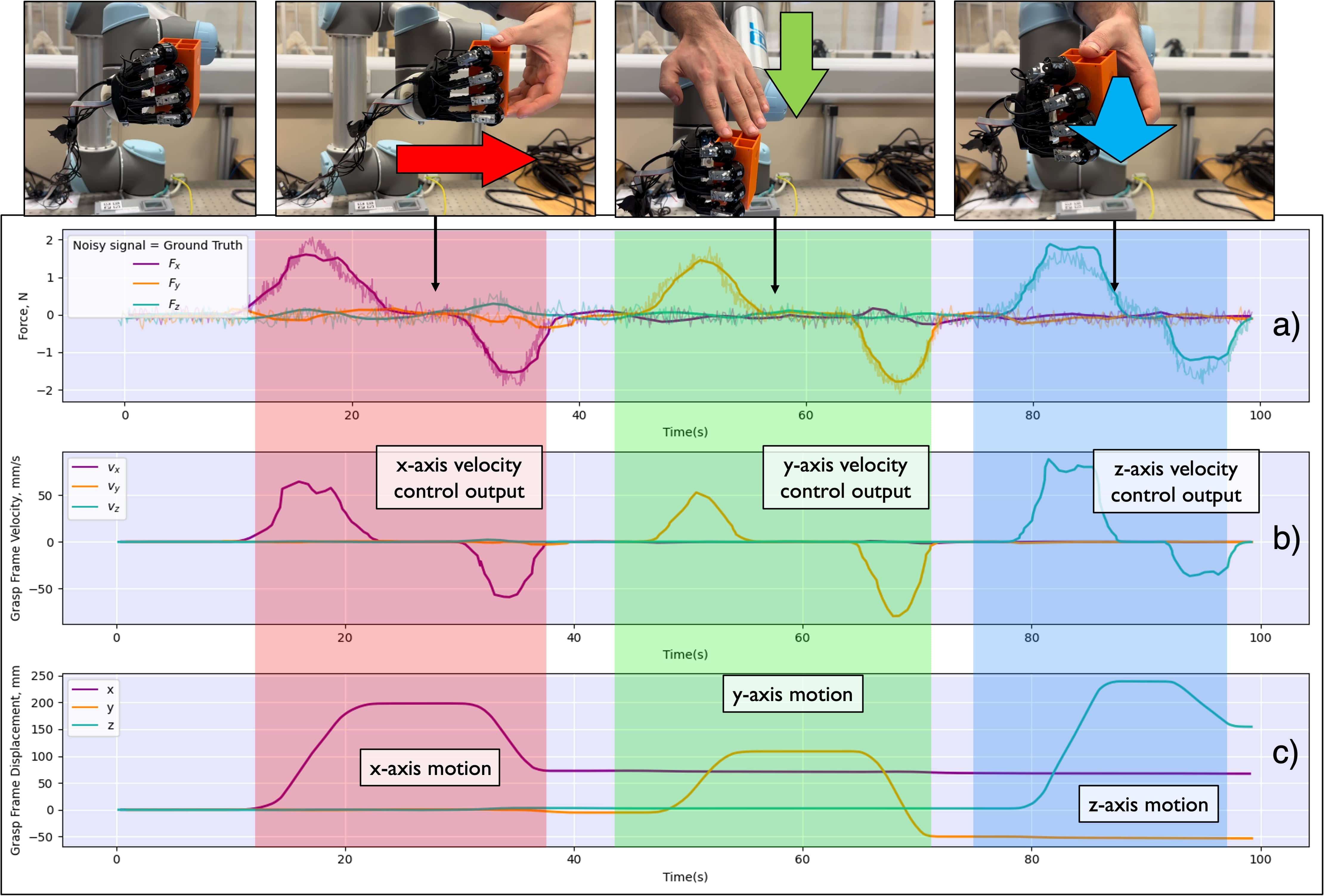}
        \vspace{-.5em}
        \caption{\textcolor{black}{Experiment 3 results: leader-follower task in which an object is grasped by the hand, then a human leader guides the object to move the hand along three axes, with the arm-hand system reacting to the force felt at the fingertips. (a) Predicted shear $F_x$, $F_y$ and normal $F_z$ forces. Ground truth values from sensorized object. (b) Resulting robot velocity signal sent to the robot arm controller, and (c) hand displacement. As the object is guided by a human hand, the trajectories are not precisely along the axes.}}
        \label{fig:exp_3_result}
\end{figure*}

\subsection{Experiment 2: Moving Robot Platform with External Disturbance}

This experiment seeks to further test the capability of the grasp controller presented in Section \ref{sec:grasp_controller} by applying it to a pouring task. This task is more complex, as the controller must account for disturbances in two directions as the mass of rice inside the cup shifts with a changing orientation to gravity in addition to the action of the rice pouring out of the cup. Again, the aim is to complete the task without crushing or dropping the cup. 

To help visualize performance, the controller variables are depicted over time (Fig.~\ref{fig:exp_2_result}) and broken down into distinct phases during the task.
\subsubsection{Initial grasp disturbance} At the outset, rice is poured into the cup in the same manner as in Experiment 1 \mbox{(Fig.~\ref{fig:exp_2_result}a-1)}. During this phase, the scaled controller inputs, $u_x$ and $u_y$, are equal to $0$ and the rate of change of shear $\Delta F_y$ is purely in the $y$-direction (Fig. \ref{fig:exp_2_result}b). These values are because the angles between the grasp frame $x$- and $y$-axes and gravity ($\theta_x$ and $\theta_y$) are $90^\circ$ and $0^\circ$ respectively (Fig. \ref{fig:exp_2_result}c). Small changes in the rate of change of shear in the $x$-direction, $\Delta F_x$, were observed as the normal force changes to compensate for the additional mass (Fig.~\ref{fig:exp_2_result}c). This affect was anticipated when designing the controller and experiment (Sec.~\ref{sec:exp_2}) and is the reason why the controller inputs are scaled according to their orientation relative to gravity. 
\subsubsection{Pouring motion begins} Once the rice has been added and the grasp stabilized, the robot begins the rotation in order to pour the rice (Fig. \ref{fig:exp_2_result}a-2), exhibited  by $\theta_x$ and $\theta_y$ decreasing and increasing respectively (Fig. \ref{fig:exp_2_result}c). This change in $\theta_x$ means that the controller input $u_x$ in the $x$-direction is non-zero and slowly increases as a scaled proportion of the rate of $x$-shear, $\Delta F_x$ (Fig. \ref{fig:exp_2_result}b). Then the dominant shear direction is along the $x$-axis, consistent with the controller behavior (Eq.~\ref{eqn:ux}). Once there has been a $90^\circ$ change of rotation (Fig. \ref{fig:exp_2_result}a-3), the $F_x$ component of the shear force is aligned to gravity, and the controller input $u_x$ becomes equal to the change in $x$-shear, $\Delta F_x$ (Fig. \ref{fig:exp_2_result}b). Conversely, when the rotation begins, the change $y$-shear, $\Delta F_y$, becomes less dominant as $\theta_y$ nears the full rotation at $90^\circ$ (Fig. \ref{fig:exp_2_result}c). Thus, the controller input $u_y$ diverges from $\Delta F_y$ (Fig. \ref{fig:exp_2_result}b) as the $y$-direction becomes less dominant up until $\theta_y = 90^\circ$, where the $y$-controller input becomes zero, $u_y = 0$. A key emergent aspect of this control is that throughout this motion, the normal force increases steadily to ensure that cup is retained in the grasp (Fig. \ref{fig:exp_2_result}b).
\subsubsection{Hand inverts to complete the pour} When the hand is upside-down, then its orientation with respect to gravity, $\theta_y > 90^\circ$ (Fig. \ref{fig:exp_2_result}b), and so $u_y$ is a scaled, inverted form of the change in $y$-shear, $\Delta F_y$ (Fig. \ref{fig:exp_2_result}a). This behavior is in accordance with the intended design of the controller (Eq.~\ref{eqn:ux}). Beyond this point, until $\theta_y = 120^\circ$, rice is poured from the cup (Fig. \ref{fig:exp_2_result}a-4). Again, a key aspect of this controller is the regulation of the normal force (Fig.~\ref{fig:exp_2_result}c), which holds the grasp steady during this phase.
\subsubsection{Motion resets} Once all the rice has been poured, the robot reverses through the same rotation trajectory. We see the orientation of the grasp frame with respect to gravity, $\theta_x$ and $\theta_y$, changes accordingly (Fig.~\ref{fig:exp_2_result}c); meanwhile, the change in shear forces $\Delta F_x$ and $\Delta F_y$ experience relatively little change due to the now much lower mass of the held object (Fig. \ref{fig:exp_2_result}b), meaning the normal force holds steady (Fig. \ref{fig:exp_2_result}c). This motion continues until the change in $y$-shear, $\Delta F_y$, starts to become dominant again and the grasp slackens, shown by a decrease in normal force. The final normal force is slightly higher than at the outset, likely due to the elastic nature of the cup. 

In total, 20 tests were run for this experiment. For every test, the cup was not crushed or dropped, showing that the controller is capable of handling delicate objects without damaging them even under external disturbances with complex dynamic conditions. The plots in Fig.~\ref{fig:exp_2_result} show a typical case. 

\subsection{Experiment 3: Tactile-Driven Leader-Follower Task}


The final experiment aims to show the capability of the shear-based grasp control in a different manner to the filling and pouring tasks, by instead converting 3D force into a velocity control signal for the UR5 robot arm, so as to enable physical human-robot interaction. The aim of this experiment is to show that the methods of deriving a velocity control signal from tactile force predictions (Sec.~\ref{control3}) allow a human user to manually apply forces to an object held by the robot, to guide it around the workspace with minimal noise or drift. 

The controller gives the desired behavior of human-guided object motion, which is illustrated by asking the user to guide the object back-and-forth along the $x$-, $y$- and $z$-axes of the workspace (Fig. \ref{fig:exp_3_result}). Even though the predicted force inputs to the controller are not always purely in the desired axis of motion (Fig. \ref{fig:exp_3_result}a), they are translated into appropriate velocity control outputs (Fig. \ref{fig:exp_3_result}b) and consequently smooth displacements in the correct direction (Fig. \ref{fig:exp_3_result}c). \textcolor{black}{The ground truth force measurements from the sensorized object (Fig.~\ref{fig:exp_3_result}a) validate that the force predictions underlying the controller are accurate. Small discrepancies, such as the overshoot on $F_z$, are likely due to the assumptions made on sensor orientation when construction the grasp frame (Section~\ref{sec:grasp_controller}).} Overall, the shear-based controller for guiding the robot under external forces (Sec.~\ref{control3}) has guided the object appropriately along the desired smooth trajectory, with noise in the force having little affect on the velocity. 


\section{Discussion}
Overall, this work introduced grasp and robot control methods for underactuated robot hands mounted on robot arms using shear and normal poses and forces felt on the tactile fingertips of the hand. This control framework was applied to several grasp-manipulation experiments: first, retaining a flexible cup in a grasp without crushing it under changes in object weight; second, a pouring task where the centre of mass of the cup changes dynamically; and third, a tactile-driven leader-follower task where a human guides a held object. 

To do these experiments, the robotic system required the integration and development of multiple robot and sensing technologies. We designed and fabricated custom soft biomimetic optical tactile sensors based on the TacTip, called microTacs, to integrate with the fingertips of an anthropomorphic soft robot hand, the Pisa/IIT SoftHand. The developed control required rapid data capture and processing from these tactile sensors, for which we developed a novel multi-input computational processing array that can capture, process and apply neural network models in parallel to tactile images from multiple sensors simultaneously at high resolution. The outputs of this array then fed into a PC that could simultaneously control the robot arm and SoftHand. As the field of tactile robotics progresses towards controlling robotic hands equipped with many high-resolution tactile sensors, we expect that hardware solutions such as this will become widely adopted because standard personal computers are not able to provide the processing and computational requirements. 

A key aspect of achieving the desired tactile robotic control was the accurate prediction of shear and normal force against the local surface of the object for each tactile fingertip. This was achieved using supervised deep learning methods based on those developed for pose-based tactile servo control with soft tactile sensors~\cite{lepora2020optimal,lepora2021pose,lloyd2021goal} that have been extended to pose and shear for a wider range of tasks~\cite{lloyd2023pose}. However, the use of five tactile sensors concurrently (rather than one previously), led to an investigation of how to utilize the training data for the best pose and force models on all tactile sensors, given that small fabrication differences meant that data could not simply be combined. We found a combination of transfer learning and individual training for each sensor gave the best models. 

Another central feature of the tactile sensing and control was the capability to quickly and accurately detect and respond to changes in shear force at the point of contact. Only by detecting shear force can the controller adjust the exerted grasp force and orientation of the robot hand. This emphasises the importance of shear force for dexterous manipulation, and a need for tactile sensors that are highly sensitive to shear. Historically, many tactile sensors have detected only normal force, but more recent tactile sensors (particularly optical tactile sensors using cameras) such as the TacTip used here are attuned in their designs to shear sensing. The importance of shear sensing in touch has been widely recognized for a while, particularly for tactile slip detection where there are many types of tactile sensor that are well suited~\cite{chen2018tactile}.

Additionally, our results validate the viewpoint that tactile feedback compensates for the lower dexterity of underactuated manipulators. Specifically, the integration of high-resolution tactile sensors gives an accurate measure of grasp force that would be difficult to determine otherwise with such a high degree of underactuation. This allows the grasp force to be carefully modulated, enabling the non-destructive handling of delicate objects under external disturbances.

In terms of limitations, the robotic system presently lacks finger joint position feedback, meaning that the exact position of the tactile sensors in space cannot be known. If this were resolved, the contact forces could be interpreted relative to the global frame more accurately, resulting in more accurate estimates of the overall poses and forces. This could be addressed by adding position sensing to the fingertips, e.g. with IMU sensors that have been previously integrated with the Pisa/IIT SoftHand~\cite{catalano2014adaptive,ajoudani2016reflex}. Furthermore, while the SoftHand benefits from the simplicity of its design that requires only one motor, having more degrees of actuation in the hand would open up a broader range of tactile manipulations tasks than we examined here. In general, we expect there will be a balance between maintaining simplicity in the control and design of the hand from having fewer actuators, and imparting additional dexterity from increasing the degrees of actuation.



For a final comment, we point out that the principles used in the design of the controller, such as changing actuator position state to maintain an equilibrium shear state, apply to a wider variety of tasks than those considered here. An interesting situation is if the methods described in this paper were to be applied to a more highly-actuated combination of robot arm and hand to enable tasks with more advanced dexterity.

\section{Conclusion}
In this paper, we presented a novel design of the TacTip optical tactile sensor that is the size of a human fingertip yet has a significantly increased extrinsic resolution compared to that of the standard design. We also developed a deep learning methodology to predict 3D pose and force, using a transfer learning architecture to improve prediction accuracy. Five such sensors were mounted as the fingertips of a Pisa/IIT SoftHand on a UR5 robot arm, which was used to perform a series of manipulation experiments. The first experiment used shear force predictions from the sensors to maintain a stable grasp on a deformable cup as mass (rice) was added without crushing the cup. The next experiment adjusted the grasp as the hand moved the cup to pour the rice out, requiring modulation of the grasp force as the mass was removed. The last experiment used translated force predictions into robot velocity to perform a tactile leader-follower task. Overall, the methods and concepts presented in this paper represent a step forward in facilitating tactile-driven manipulation and therefore more human-like dexterity in robots. 
\bibliographystyle{IEEEtran}
\bibliography{IEEEabrv,ref}

\begin{thebibliography}{10}
\providecommand{\url}[1]{#1}
\csname url@samestyle\endcsname
\providecommand{\newblock}{\relax}
\providecommand{\bibinfo}[2]{#2}
\providecommand{\BIBentrySTDinterwordspacing}{\spaceskip=0pt\relax}
\providecommand{\BIBentryALTinterwordstretchfactor}{4}
\providecommand{\BIBentryALTinterwordspacing}{\spaceskip=\fontdimen2\font plus
\BIBentryALTinterwordstretchfactor\fontdimen3\font minus \fontdimen4\font\relax}
\providecommand{\BIBforeignlanguage}[2]{{%
\expandafter\ifx\csname l@#1\endcsname\relax
\typeout{** WARNING: IEEEtran.bst: No hyphenation pattern has been}%
\typeout{** loaded for the language `#1'. Using the pattern for}%
\typeout{** the default language instead.}%
\else
\language=\csname l@#1\endcsname
\fi
#2}}
\providecommand{\BIBdecl}{\relax}
\BIBdecl

\bibitem{bicchi2000robotic}
A.~Bicchi and V.~Kumar, ``Robotic grasping and contact: A review,'' in \emph{Proceedings 2000 ICRA. Millennium conference. IEEE international conference on robotics and automation. Symposia proceedings (Cat. No. 00CH37065)}, vol.~1, 2000, pp. 348--353.

\bibitem{birglen2004kinetostatic}
L.~Birglen and C.~M. Gosselin, ``Kinetostatic analysis of underactuated fingers,'' \emph{IEEE Transactions on Robotics and Automation}, vol.~20, no.~2, pp. 211--221, 2004.

\bibitem{okamura2000overview}
A.~M. Okamura, N.~Smaby, and M.~R. Cutkosky, ``An overview of dexterous manipulation,'' in \emph{Proceedings 2000 ICRA. Millennium Conference. IEEE International Conference on Robotics and Automation. Symposia Proceedings (Cat. No. 00CH37065)}, vol.~1, 2000, pp. 255--262.

\bibitem{10161036}
C.~J. Ford, H.~Li, J.~Lloyd, M.~G. Catalano, M.~Bianchi, E.~Psomopoulou, and N.~F. Lepora, ``Tactile-driven gentle grasping for human-robot collaborative tasks,'' in \emph{2023 IEEE International Conference on Robotics and Automation (ICRA)}, 2023, pp. 10\,394--10\,400.

\bibitem{lloyd2023pose}
J.~Lloyd and N.~F. Lepora, ``Pose-and-shear-based tactile servoing,'' \emph{The International Journal of Robotics Research}, vol.~43, no.~7, pp. 1024--1055, 2024.

\bibitem{kappassov2015tactile}
Z.~Kappassov, J.-A. Corrales, and V.~Perdereau, ``Tactile sensing in dexterous robot hands,'' \emph{Robotics and Autonomous Systems}, vol.~74, pp. 195--220, 2015.

\bibitem{james2018slip}
J.~W. James, N.~Pestell, and N.~F. Lepora, ``Slip detection with a biomimetic tactile sensor,'' \emph{IEEE Robotics and Automation Letters}, vol.~3, no.~4, pp. 3340--3346, 2018.

\bibitem{james2020slip}
J.~W. James and N.~F. Lepora, ``Slip detection for grasp stabilization with a multifingered tactile robot hand,'' \emph{IEEE Transactions on Robotics}, vol.~37, no.~2, pp. 506--519, 2020.

\bibitem{lepora2021soft}
N.~F. Lepora, ``Soft biomimetic optical tactile sensing with the {TacTip}: A review,'' \emph{IEEE Sensors Journal}, vol.~21, no.~19, pp. 21\,131--21\,143, 2021.

\bibitem{ward2018tactip}
B.~Ward-Cherrier, N.~Pestell, L.~Cramphorn, B.~Winstone, M.~E. Giannaccini, J.~Rossiter, and N.~F. Lepora, ``The {TacTip} family: Soft optical tactile sensors with 3d-printed biomimetic morphologies,'' \emph{Soft robotics}, vol.~5, no.~2, pp. 216--227, 2018.

\bibitem{bulens2023incipient}
D.~C. Bulens, N.~F. Lepora, S.~J. Redmond, and B.~Ward-Cherrier, ``Incipient slip detection with a biomimetic skin morphology,'' in \emph{2023 IEEE/RSJ International Conference on Intelligent Robots and Systems (IROS)}, 2023, pp. 8972--8978.

\bibitem{chen2018tactile}
W.~Chen, H.~Khamis, I.~Birznieks, N.~F. Lepora, and S.~J. Redmond, ``Tactile sensors for friction estimation and incipient slip detection—toward dexterous robotic manipulation: A review,'' \emph{IEEE Sensors Journal}, vol.~18, no.~22, pp. 9049--9064, 2018.

\bibitem{catalano2014adaptive}
M.~G. Catalano, G.~Grioli, E.~Farnioli, A.~Serio, C.~Piazza, and A.~Bicchi, ``Adaptive synergies for the design and control of the {Pisa/IIT} softhand,'' \emph{The International Journal of Robotics Research}, vol.~33, no.~5, pp. 768--782, 2014.

\bibitem{lepora2021towards}
N.~F. Lepora, C.~Ford, A.~Stinchcombe, A.~Brown, J.~Lloyd, M.~G. Catalano, M.~Bianchi, and B.~Ward-Cherrier, ``Towards integrated tactile sensorimotor control in anthropomorphic soft robotic hands,'' in \emph{2021 IEEE International Conference on Robotics and Automation (ICRA)}, 2021, pp. 1622--1628.

\bibitem{howe1993tactile}
R.~D. Howe, ``Tactile sensing and control of robotic manipulation,'' \emph{Advanced Robotics}, vol.~8, no.~3, pp. 245--261, 1993.

\bibitem{choi2012external}
J.~Choi, S.~Kang \emph{et~al.}, ``External force estimation using joint torque sensors for a robot manipulator,'' in \emph{2012 IEEE International Conference on Robotics and Automation}, 2012, pp. 4507--4512.

\bibitem{daoud2012real}
N.~Daoud, J.-P. Gazeau, S.~Zeghloul, and M.~Arsicault, ``A real-time strategy for dexterous manipulation: Fingertips motion planning, force sensing and grasp stability,'' \emph{Robotics and Autonomous Systems}, vol.~60, no.~3, pp. 377--386, 2012.

\bibitem{nguyen2013fingertip}
K.-C. Nguyen and V.~Perdereau, ``Fingertip force control based on max torque adjustment for dexterous manipulation of an anthropomorphic hand,'' in \emph{2013 IEEE/RSJ international conference on intelligent robots and systems}, 2013, pp. 3557--3563.

\bibitem{liu2015finger}
H.~Liu, K.~C. Nguyen, V.~Perdereau, J.~Bimbo, J.~Back, M.~Godden, L.~D. Seneviratne, and K.~Althoefer, ``Finger contact sensing and the application in dexterous hand manipulation,'' \emph{Autonomous Robots}, vol.~39, pp. 25--41, 2015.

\bibitem{grioli2012adaptive}
G.~Grioli, M.~Catalano, E.~Silvestro, S.~Tono, and A.~Bicchi, ``Adaptive synergies: an approach to the design of under-actuated robotic hands,'' in \emph{2012 IEEE/RSJ International Conference on Intelligent Robots and Systems}, 2012, pp. 1251--1256.

\bibitem{barkat2009optimization}
B.~Barkat, S.~Zeghloul, and J.-P. Gazeau, ``Optimization of grasping forces in handling of brittle objects,'' \emph{Robotics and Autonomous Systems}, vol.~57, no.~4, pp. 460--468, 2009.

\bibitem{xu2021compliant}
W.~Xu, H.~Zhang, H.~Yuan, and B.~Liang, ``A compliant adaptive gripper and its intrinsic force sensing method,'' \emph{IEEE Transactions on Robotics}, vol.~37, no.~5, pp. 1584--1603, 2021.

\bibitem{tegin2005tactile}
J.~Tegin and J.~Wikander, ``Tactile sensing in intelligent robotic manipulation--a review,'' \emph{Industrial Robot: An International Journal}, vol.~32, no.~1, pp. 64--70, 2005.

\bibitem{yousef2011tactile}
H.~Yousef, M.~Boukallel, and K.~Althoefer, ``Tactile sensing for dexterous in-hand manipulation in robotics—a review,'' \emph{Sensors and Actuators A: physical}, vol. 167, no.~2, pp. 171--187, 2011.

\bibitem{ajoudani2016reflex}
A.~Ajoudani, E.~Hocaoglu, A.~Altobelli, M.~Rossi, E.~Battaglia, N.~Tsagarakis, and A.~Bicchi, ``Reflex control of the {Pisa/IIT} softhand during object slippage,'' in \emph{2016 IEEE International Conference on Robotics and Automation (ICRA)}, 2016, pp. 1972--1979.

\bibitem{battaglia2015thimblesense}
E.~Battaglia, M.~Bianchi, A.~Altobelli, G.~Grioli, M.~G. Catalano, A.~Serio, M.~Santello, and A.~Bicchi, ``Thimblesense: a fingertip-wearable tactile sensor for grasp analysis,'' \emph{IEEE transactions on haptics}, vol.~9, no.~1, pp. 121--133, 2015.

\bibitem{ATI_Nano17}
\BIBentryALTinterwordspacing
{ATI Industrial Automation}, ``Ati force/torque sensor - nano17,'' 2024, accessed: 2024-08-20. [Online]. Available: \url{https://www.ati-ia.com/products/ft/ft_models.aspx?id=Nano17}
\BIBentrySTDinterwordspacing

\bibitem{li2013control}
Q.~Li, C.~Sch{\"u}rmann, R.~Haschke, and H.~J. Ritter, ``A control framework for tactile servoing.'' in \emph{Robotics: Science and systems}, 2013.

\bibitem{kappassov2020touch}
Z.~Kappassov, J.-A. Corrales, and V.~Perdereau, ``Touch driven controller and tactile features for physical interactions,'' \emph{Robotics and Autonomous Systems}, vol. 123, p. 103332, 2020.

\bibitem{lepora2021pose}
N.~F. Lepora and J.~Lloyd, ``Pose-based tactile servoing: Controlled soft touch using deep learning,'' \emph{IEEE Robotics \& Automation Magazine}, vol.~28, no.~4, pp. 43--55, 2021.

\bibitem{lepora2020optimal}
------, ``Optimal deep learning for robot touch: Training accurate pose models of 3d surfaces and edges,'' \emph{IEEE Robotics \& Automation Magazine}, vol.~27, no.~2, pp. 66--77, 2020.

\bibitem{de2017multimodal}
T.~E.~A. De~Oliveira, A.-M. Cretu, and E.~M. Petriu, ``Multimodal bio-inspired tactile sensing module,'' \emph{IEEE Sensors Journal}, vol.~17, no.~11, pp. 3231--3243, 2017.

\bibitem{lepora2019pixels}
N.~F. Lepora, A.~Church, C.~De~Kerckhove, R.~Hadsell, and J.~Lloyd, ``From pixels to percepts: Highly robust edge perception and contour following using deep learning and an optical biomimetic tactile sensor,'' \emph{IEEE Robotics and Automation Letters}, vol.~4, no.~2, pp. 2101--2107, 2019.

\bibitem{lloyd2021goal}
J.~Lloyd and N.~F. Lepora, ``Goal-driven robotic pushing using tactile and proprioceptive feedback,'' \emph{IEEE Transactions on Robotics}, vol.~38, no.~2, pp. 1201--1212, 2021.

\bibitem{andrychowicz2020learning}
O.~M. Andrychowicz, B.~Baker, M.~Chociej, R.~Jozefowicz, B.~McGrew, J.~Pachocki, A.~Petron, M.~Plappert, G.~Powell, A.~Ray \emph{et~al.}, ``Learning dexterous in-hand manipulation,'' \emph{The International Journal of Robotics Research}, vol.~39, no.~1, pp. 3--20, 2020.

\bibitem{cabas2006optimized}
R.~Cab{\'a}s, L.~M. Cabas, and C.~Balaguer, ``Optimized design of the underactuated robotic hand,'' in \emph{Proceedings 2006 IEEE International Conference on Robotics and Automation, 2006. ICRA 2006.}, 2006, pp. 982--987.

\bibitem{deimel2016novel}
R.~Deimel and O.~Brock, ``A novel type of compliant and underactuated robotic hand for dexterous grasping,'' \emph{The International Journal of Robotics Research}, vol.~35, no. 1-3, pp. 161--185, 2016.

\bibitem{li2022brl}
H.~Li, C.~J. Ford, M.~Bianchi, M.~G. Catalano, E.~Psomopoulou, and N.~F. Lepora, ``{BRL/Pisa/IIT} {SoftHand}: A low-cost, 3d-printed, underactuated, tendon-driven hand with soft and adaptive synergies,'' \emph{IEEE Robotics and Automation Letters}, vol.~7, no.~4, pp. 8745--8751, 2022.

\bibitem{shintake2018soft}
J.~Shintake, V.~Cacucciolo, D.~Floreano, and H.~Shea, ``Soft robotic grippers,'' \emph{Advanced materials}, vol.~30, no.~29, p. 1707035, 2018.

\bibitem{niehues2015compliance}
T.~D. Niehues, P.~Rao, and A.~D. Deshpande, ``Compliance in parallel to actuators for improving stability of robotic hands during grasping and manipulation,'' \emph{The International Journal of Robotics Research}, vol.~34, no.~3, pp. 256--269, 2015.

\bibitem{della2018toward}
C.~Della~Santina, C.~Piazza, G.~Grioli, M.~G. Catalano, and A.~Bicchi, ``Toward dexterous manipulation with augmented adaptive synergies: The {Pisa/IIT} softhand 2,'' \emph{IEEE Transactions on Robotics}, vol.~34, no.~5, pp. 1141--1156, 2018.

\bibitem{li2023marker}
M.~Li, T.~Li, and Y.~Jiang, ``Marker displacement method used in vision-based tactile sensors—from 2d to 3d-a review,'' \emph{IEEE Sensors Journal}, 2023.

\bibitem{wang2021gelsight}
S.~Wang, Y.~She, B.~Romero, and E.~Adelson, ``{GelSight} wedge: Measuring high-resolution 3d contact geometry with a compact robot finger,'' in \emph{2021 IEEE International Conference on Robotics and Automation (ICRA)}, 2021, pp. 6468--6475.

\bibitem{lepora2022digitac}
N.~F. Lepora, Y.~Lin, B.~Money-Coomes, and J.~Lloyd, ``Digitac: A {DIGIT-TacTip} hybrid tactile sensor for comparing low-cost high-resolution robot touch,'' \emph{IEEE Robotics and Automation Letters}, vol.~7, no.~4, pp. 9382--9388, 2022.

\bibitem{lu2024dexitac}
C.~Lu, K.~Tang, M.~Yang, T.~Yue, H.~Li, and N.~F. Lepora, ``{DexiTac}: Soft dexterous tactile gripping,'' \emph{IEEE/ASME Transactions on Mechatronics}, vol.~30, no.~1, pp. 333--344, 2025.

\bibitem{yang2024anyrotate}
M.~Yang, C.~Lu, A.~Church, Y.~Lin, C.~Ford, H.~Li, E.~Psomopoulou, D.~A. Barton, and N.~F. Lepora, ``{AnyRotate}: Gravity-invariant in-hand object rotation with sim-to-real touch,'' 2024.

\bibitem{lambeta2020digit}
M.~Lambeta, P.-W. Chou, S.~Tian, B.~Yang, B.~Maloon, V.~R. Most, D.~Stroud, R.~Santos, A.~Byagowi, G.~Kammerer \emph{et~al.}, ``{DIGIT}: A novel design for a low-cost compact high-resolution tactile sensor with application to in-hand manipulation,'' \emph{IEEE Robotics and Automation Letters}, vol.~5, no.~3, pp. 3838--3845, 2020.

\bibitem{abad2020visuotactile}
A.~C. Abad and A.~Ranasinghe, ``Visuotactile sensors with emphasis on {GelSight} sensor: A review,'' \emph{IEEE Sensors Journal}, vol.~20, no.~14, pp. 7628--7638, 2020.

\bibitem{cramphorn2017addition}
L.~Cramphorn, B.~Ward-Cherrier, and N.~F. Lepora, ``Addition of a biomimetic fingerprint on an artificial fingertip enhances tactile spatial acuity,'' \emph{IEEE Robotics and Automation Letters}, vol.~2, no.~3, pp. 1336--1343, 2017.

\bibitem{shi2012application}
Y.~Shi, ``The application of the butterworth low-pass digital filter on experimental data processing,'' in \emph{2011 International Conference in Electrics, Communication and Automatic Control Proceedings}.\hskip 1em plus 0.5em minus 0.4em\relax Springer, 2012, pp. 225--230.

\bibitem{kurapa2020hybrid}
A.~Kurapa, D.~Rathore, D.~R. Edla, A.~Bablani, and V.~Kuppili, ``A hybrid approach for extracting emg signals by filtering eeg data for iot applications for immobile persons,'' \emph{Wireless Personal Communications}, vol. 114, pp. 3081--3101, 2020.

\bibitem{yu2022progressive}
Z.~Yu, D.~Shen, Z.~Jin, J.~Huang, D.~Cai, and X.-S. Hua, ``Progressive transfer learning,'' \emph{IEEE Transactions on Image Processing}, vol.~31, pp. 1340--1348, 2022.

\bibitem{zhuang2020comprehensive}
F.~Zhuang, Z.~Qi, K.~Duan, D.~Xi, Y.~Zhu, H.~Zhu, H.~Xiong, and Q.~He, ``A comprehensive survey on transfer learning,'' \emph{Proceedings of the IEEE}, vol. 109, no.~1, pp. 43--76, 2020.

\bibitem{wang2004image}
Z.~Wang, A.~C. Bovik, H.~R. Sheikh, and E.~P. Simoncelli, ``Image quality assessment: from error visibility to structural similarity,'' \emph{IEEE transactions on image processing}, vol.~13, no.~4, pp. 600--612, 2004.

\bibitem{james2021tactile}
J.~W. James, A.~Church, L.~Cramphorn, and N.~F. Lepora, ``Tactile {Model} {O}: Fabrication and testing of a 3d-printed, three-fingered tactile robot hand,'' \emph{Soft Robotics}, vol.~8, no.~5, pp. 594--610, 2021.

\bibitem{sun2017revisiting}
C.~Sun, A.~Shrivastava, S.~Singh, and A.~Gupta, ``Revisiting unreasonable effectiveness of data in deep learning era,'' in \emph{Proceedings of the IEEE international conference on computer vision}, 2017, pp. 843--852.

\bibitem{peng2019domain}
X.~Peng, Z.~Huang, X.~Sun, and K.~Saenko, ``Domain agnostic learning with disentangled representations,'' in \emph{International conference on machine learning}.\hskip 1em plus 0.5em minus 0.4em\relax PMLR, 2019, pp. 5102--5112.

\end{thebibliography}

\appendices
\section{Learning Parameters} \label{appA}
\begin{table}[h!]
\centering
\caption{Network hyperparameters}
    \arrayrulecolor{black}
    \begin{tabular}{|l|l|} 
    \hline
    \textbf{Parameter}                    & \textbf{Value}   \\ \hline
    Number of convolutional hidden layers & 4                \\
    Number of convolutional kernels       & 480              \\
    Number of fully connected layers      & 2                \\
    Number of dense hidden layer units    & 512              \\
    Hidden layer activation function      & ReLU             \\
    Dropout coefficient                   & 0                \\
    Batch size                            & 16               \\
    Adam decay                            & $1\times10^{-6}$ \\
    Adam $\beta_1$                        & 0.9              \\
    Adam $\beta_2$                         & 0.999            \\ \hline
    \end{tabular} \label{tab:hyperparams}
\end{table}

Each convolutional layer used a 3x3 kernel with unit stride and no padding, applying batch normalisation followed by a max pooling operation to reduce the chance of overfitting. The output of each layer is then passed through a ReLU activation function, with the output of the final convolutional layer then passed through a series of fully connected layers before being combined in a linear output layer to give a vector of pose and force predictions. The weights and biases of the model were optimised by minimising the mean squared error (MSE) during training and the Adam optimiser was used to apply an adaptive learning rate. Training code was written in PyTorch and executed via CUDA on an Nvidia RTX 2060 GPU.

\section{Scaling} \label{appB}
The scaling functions described in section \ref{sec:exp_2} are derived from the trigonometric expression for a periodic triangular cosine wave with amplitude $a$ and period $p$:
\begin{equation}
    \label{eq:triangle}
    f(\theta) = 1- \frac{2a}{\pi}\arccos\left ( \cos\left ( \frac{2\pi}{p}\theta \right ) \right ).
\end{equation}
To achieve the behavior previously described, $a=1$ such that the output is scaled between -1 and 1 and $p=2\pi$, meaning the output is 0 when $\theta=90^\circ$ (i.e. when perpendicular to gravity). Substituting these values into Equation \ref{eq:triangle} gives:
\begin{equation}
    \label{eqn:s}
    f(\theta) = 1 - \frac{2\theta}{\pi}. 
\end{equation}
Equations \ref{eqn:sx} are then found by substituting $\theta_x$ and $\theta_y$ for $\theta$ in Equation \ref{eqn:s}. Lastly, Equations \ref{eqn:ux} are then given by multiplying Equations \ref{eqn:sx} by $\Delta F_x$ and $\Delta F_y$ respectively.

\section{Control Parameters} \label{appC}

\begin{table}[h!]
\centering
\caption{Table of Control Parameters}
\arrayrulecolor{black}
\begin{tabular}{|cc||cc||cc|}
\hline
\rowcolor[HTML]{C0C0C0} 
\multicolumn{2}{|c||}{\cellcolor[HTML]{C0C0C0}\textbf{Experiment 1}}          & \multicolumn{2}{c||}{\cellcolor[HTML]{C0C0C0}\textbf{Experiment 2}}          & \multicolumn{2}{c|}{\cellcolor[HTML]{C0C0C0}\textbf{Experiment 3}}          \\ \hline
\rowcolor[HTML]{EFEFEF} 
\multicolumn{1}{|c|}{\cellcolor[HTML]{EFEFEF}\textbf{Gain}} & \textbf{Value} & \multicolumn{1}{c|}{\cellcolor[HTML]{EFEFEF}\textbf{Gain}} & \textbf{Value} & \multicolumn{1}{c|}{\cellcolor[HTML]{EFEFEF}\textbf{Gain}} & \textbf{Value} \\ \hline
\multicolumn{1}{|c|}{$K_P$}                                 & 110            & \multicolumn{1}{c|}{$K_{Px}$}                              & 300            & \multicolumn{1}{c|}{$K_{0,x}$}                             & 500            \\
\multicolumn{1}{|c|}{$K_I$}                                 & 0.05           & \multicolumn{1}{c|}{$K_{Ix}$}                              & 1              & \multicolumn{1}{c|}{$K_{0,y}$}                             & 500            \\
\multicolumn{1}{|c|}{$K_D$}                                 & 0.01           & \multicolumn{1}{c|}{$K_{Dx}$}                              & 0.1            & \multicolumn{1}{c|}{$K_{0,z}$}                             & 1500           \\
\multicolumn{1}{|c|}{-}                                     & -              & \multicolumn{1}{c|}{$K_{Py}$}                              & 110            & \multicolumn{1}{c|}{-}                                     & -              \\
\multicolumn{1}{|c|}{-}                                     & -              & \multicolumn{1}{c|}{$K_{Iy}$}                              & 0.05           & \multicolumn{1}{c|}{-}                                     & -              \\
\multicolumn{1}{|c|}{-}                                     & -              & \multicolumn{1}{c|}{$K_{Dy}$}                              & 0.01           & \multicolumn{1}{c|}{-}                                     & -              \\ \hline
\end{tabular} \label{tab:control_params}
\end{table}

\end{document}